\documentclass[final,3p,times,onecolumn]{elsarticle}

\usepackage[section]{placeins}
\usepackage{color}
\usepackage{float}
\usepackage{epsfig}
\usepackage{graphicx, subfigure}
\usepackage{amssymb, url}
\usepackage{array}
\usepackage{tabularx}
\usepackage{rotating}
\usepackage{amsmath}
\usepackage{algorithm}  
\usepackage{algorithmic}  
\usepackage{hyperref}
\usepackage{natbib}
\usepackage{multirow}
\usepackage{indentfirst}
\usepackage[justification=centering]{caption}
\definecolor{mygreen}{rgb}{0,0.6,0}
\definecolor{mygray}{rgb}{0.5,0.5,0.5}
\definecolor{mymauve}{rgb}{0.58,0,0.82}

\newcommand{\tabincell}[2]{\begin{tabular}{@{}#1@{}}#2\end{tabular}}
\newcommand{\sys}{M\&M}

\begin{document}
\begin{frontmatter}

\title{Towards Explainable Meta-Learning for DDoS Detection}

\author[cs]{Qianru~Zhou}
\author[cs]{Rongzhen~Li}
\author[cs]{Lei~Xu\corref{cor}}
\author[uk]{Arumugam Nallanathan}
\author[cs]{Jian~Yang}
\author[cs]{Anmin~Fu}
\cortext[cor]{Corresponding author. zhouqianru@njust.edu.cn}
\address[cs]{School of Compute Science and Engineering, Nanjing University of Science and Technology, Nanjing, China.}
\address[uk]{School of Electronic Engineering and Computer Science, Queen Mary University of London, London E1 4NS, U.K.}

\begin{abstract}
The Internet is the most complex machine humankind has ever built, and how to immune it from intrusions is even more complex. With the ever increasing of new intrusions, intrusion detection tasks are increasingly rely on Artificial Intelligence. Interpretability and transparency of the machine learning model is the foundation of trust in AI-driven intrusion detection. Current interpretable Artificial Intelligence technologies in intrusion detection are heuristic, which is neither accurate nor sufficient. This paper proposed a rigorous interpretable Artificial Intelligence driven intrusion detection approach, based on artificial immune system. Details of rigorous interpretation calculation process for the decision tree model are presented. A map, combine and merge (M\&M) method is proposed to discretize continuous features into boolean expression and simplify into prime implicants. \emph{Prime implicant} explanations for DDoS LOIC and HOIC attack traffic flows are given in detail as rules for the DDoS intrusion detection system. Experiments are carried out use real-life traffic to evaluate the system, it is evident that as the interpretable method is based on formal logic calculation process, the explanation provides rigorous and sufficient reasons for the decision of DDoS traffic flow classification.
\end{abstract}

\begin{keyword}
Interpretable machine learning, explainable Artificial Intelligence, prime implicant, intrusion detection system, DDoS attack.
\end{keyword}
\end{frontmatter}


\section{Introduction}

Artificial Intelligent based decision making have been used broadly in various domains across industries, government, and everyday life, such as facial recognition, loan assessment, bail assessment, healthcare, self-driving vehicle, and cybersecurity etc 
\cite{ribeiro2016should, ribeiro2018anchors, marino2018adversarial, guo2018lemna, kim2016examples, johansson2004truth, mahdavifar2020dennes, rago2018argumentation, brarda2020using, grover2019beef, lundberg2017unified, lundberg2020local, shakarian2016belief, darwiche2020reasons, ignatiev2019abduction}. However, in domains where security is of utmost importance, such as cybersecurity, trust is the fundamental basis and guarantee of the validity and prosperity of AI-based decision making. People's trust on the decisions made is based on the interpretability and transparency of the machine learning models make them \cite{gunning2019darpa}. 
Unfortunately, most of the popular machine learning models, such as deep learning, neural networks, and even the tree-based models are uninterpretable (although the tree-based models are believed to be interpretable for they can provide the decision paths that lead to the decisions, many have point out that these explanations are ``shallow'' and contain potentially too many redundant features and rules, and thus actually unable to provide rigorous sufficient reasons, also known as \emph{prime implicant explanations}, or \emph{minimal sufficient reasons} \cite{DBLP2021}). Consequences of the decision made by uninterpretable machine learning models are occasionally catastrophic, for example the fatal car crushes by Google's autonomous car \cite{mathur2015google} and Tesla's autopilot system \cite{banks2018driver}; An automatic bail risk assessment algorithm is believed to be biased and keep many people in jail longer than they should without explicit reasons, and another machine learning based DNA trace analysis software accuses people with crimes they did not commit \footnote{See \url{https://www.nytimes.com/2017/06/13/opinion/how-computers-are-harming-criminal-justice.html}}; Millions of African-American could not get due medical care by a biased machine learning assessment algorithm \footnote{See \url{https://www.wsj.com/articles/researchers}\\\url{-find-racial-bias-in-hospital-algorithm-11571941096}}; In Scotland, a football game is ruined because the AI camera mistakes the judge's bald head as the ball and keep focusing on it rather than the goal scene\footnote{see \url{https://www.ndtv.com/offbeat/ai-camera-ruins-}\\\url{football-game-by-mistaking-referees-bald-head-}\\\url{for-ball-2319171}}. The key reasons lay in that all machine learning models suffer from overfitting \cite{domingos2015master}. Overfitting could be seriously exacerbated by noisy data, and real-life data is, and almost always, noisy. These, among many other reasons (like GDPR requirements \cite{edwards2017slave} and judicial requirements \cite{deeks2019judicial}), have driven the surge of research interest on the interpretation of machine learning models, analyzing the reasons for positive or negative decisions, interrogating them by human domain experts, and adjusting them if necessary. That gives rise to the surge of research interest in Explainable Artificial Intelligence (XAI) or Interpretable Machine Learning (IML)\footnote{there is subtle difference between explainable and interpretable AI, but this is not within the focus of this paper, so we will use XAI to represent both methodologies throughout the paper.}\cite{ribeiro2016should, ribeiro2018anchors, marino2018adversarial, guo2018lemna, kim2016examples, johansson2004truth, mahdavifar2020dennes, rago2018argumentation, brarda2020using, grover2019beef, lundberg2017unified, lundberg2020local, shakarian2016belief, darwiche2020reasons, ignatiev2019abduction}.

In this paper, we make a modest step towards a rigorous XAI driven intrusion detection system for DDoS attacks. A map, combine and merge (\sys) methodology is proposed to transform any decision tree model with continuous features into boolean expression with formal logic calculation. By calculating the \emph{prime implicants} of the boolean expression, rigorous explanation of the model is extracted and interpreted with human natural language. A decision tree model achieved almost 100\% accuracy in training with CIC-AWS-2018 DDoS datasets is used as a demo in this paper, rigorous \sys~transformation is applied and exhaustive human-readable explanations are presented in detail.

The paper is organized as follows, Section \ref{sec_survey} provide a overall literature review for XAI methodologies and its applications in intrusion detection; Section \ref{sec_rin} provides details of proposed rigorous XAI methodologies, including continuous features discretization and prime implicants calculation, at the end of the section, proposed \sys~system is presented with its key components and architecture; Section \ref{sec_res} present the details of \emph{prime implicant} explanation calculated from the target model, together with the evaluation results tested on real-life traffic flow instances. Section \ref{sec_sum} summarize the work and discuss about a few future challenges that need to be done.

\section{Explainable Artificial Intelligent driven intrusion detections}\label{sec_survey}
\subsection{Explainable Artificial Intelligence}

\begin{table*}[htbp]
\scriptsize
\centering
\caption{Summary of XAI methodologies.}
\label{table_sum}
{
\begin{tabular}{|p{1.5cm}|p{1.3cm}|c|c|c|}
\hline
&Name & Description & Target Models & Ref. \\ \hline
 \multirow{13}*{Heuristic}  & LIME & \tabincell{c}{generate an explanation by approximating \\the underlying model by an interpretable one\\ (such as a linear model with only \\a few non-zero coefficients), learned on perturbations of the original instance \\ (e.g., removing words or hiding parts of the image)} & DNN & \cite{ribeiro2016should} \\  \cline{2-5}

& ANCHOR &  \tabincell{c}{Improved based on LIME, \\provide rule-based, model-agnostic explanations on local behaviors of the models.} & Any & \cite{ribeiro2018anchors} \\  \cline{2-5}

& \tabincell{c}{Marino \\ et.al.'s work} & \tabincell{c}{use adversarial machine learning to find the minimum \\modifications  (of the input features) required to correctly \\classify a given set of misclassified samples, \\by finding an adversarial sample that is classified as positive while minimizing \\ the distance between the real sample and the modified sample.} & DNN & \cite{marino2018adversarial} \\  \cline{2-5}

& LEMNA & \tabincell{c}{LEMNA generates a small set of interpretable features to \\ explain the model by approximating a local area \\ of the complex deep learning decision boundary.} & DNN & \cite{guo2018lemna} \\  \cline{2-5}

& MMD-critic & \tabincell{c}{Use maximum mean discrepancy (MMD), \\ a measure of the difference between distributions, \\to efficiently learns prototypes and criticism for Bayesian models.} & Bayesian & \cite{kim2016examples} \\  \cline{2-5}

& G-REX & \tabincell{c}{use a genetic rule extract method G-REX to interpret \\the knowledge represented by the architecture and the weights \\of machine learning model, generating in forms of regression trees, fuzzy rules, \\Boolean rules, and decision trees.} & Any & \cite{johansson2004truth}  \\  \cline{2-5}

&DeNNeS & \tabincell{c}{An embedded, deep learning-based cybersecurity \\ expert system extracting refined rules from \\a trained multilayer DNN, could either function on a complete or \\ incomplete dataset.} & DNN & \cite{mahdavifar2020dennes} \\  \cline{2-5}

& \tabincell{c}{TFs} & \tabincell{c}{use user-tailored Tripolar Argumentation Frameworks (TFs) \\to give explanations that can help elicit users' feedback leading to \\positive effects on the quality of future recommendations.}  & A new model & \cite{rago2018argumentation} \\ \cline{2-5}

&\tabincell{c}{Johannson \\ et.al.'s work} & \tabincell{c}{use user provided rules, perform argument-based \\reasoning mechanism in prolog, and compute the Explanation \\graphs to explain the decisions made.}  & decision system & \cite{brarda2020using}  \\ \cline{2-5}

& BEEF & \tabincell{c}{clustering algorithm over the entire dataset, \\in order to generate a local explanation \\(in both supporting and counterfactual rules).} & decision system & \cite{grover2019beef} \\  \cline{2-5}

& SHAP & \tabincell{c}{to calculate an additive feature \\ importance score for each particular prediction with a set \\of desirable properties \\(local accuracy, missingness and consistency) that its antecedents lacked.} & Any & \cite{lundberg2017unified} \\  \cline{2-5}

&\tabincell{c}{Tree \\explainer} & \tabincell{c}{compute Shapley value explanations to \\directly capture feature interactions and \\then get local explanations. By combining \\many local explanations, get an understanding of the global model structure.}  & Trees based & \cite{lundberg2020local} \\ \cline{2-5}

& DeLP3E  & \tabincell{c}{an extension of the PreDeLP probability \\reasoning language in which sentences can be \\annotated with probabilistic events. \\Use attribution query to figure out \\attributing responsibility to entities given a cyber event, \\and achieve interpretation.} & Bayesian & \cite{shakarian2016belief} \\ \hline

\multirow{2}*{rigorous}& \tabincell{c}{knowledge \\compilation} & \tabincell{c}{compile model into boolean circuits \\in the forms of CNF or DNF, and interpret the model \\by solving the prime implicant of the circuit.} & \tabincell{c}{Decision Tree \\Bayesian \\ Binary Neural Network} & \cite{darwiche2020reasons} \\ \cline{2-5}

&\tabincell{c}{Abductive\\ based \\explanation} & \tabincell{c}{represent a model into a formalized constraints\\ and provide cardinality-minimal explanations by \\applying abductive reasoning on the model to \\answer entailment queries.} & Any & \cite{ignatiev2019abduction} \\

\hline

\end{tabular}
}
\end{table*}

Started almost three decades ago, the research on Explainable AI or interpretable AI methodologies have been numerous, the methodologies proposed can be roughly classified into two types: rigorous methods (or logical reasoning methods) and heuristic methods. Most of the proposed explanation approaches are heuristic approaches, such as Shaply additive explanations (SHAP), Local Interpretable Model-Agnostic Explanations (LIME), and ANCHOR. The major issues of heuristic approaches are that they are model-agnostic and cannot guarantee accuracy, thus, they are not really trustworthy. Besides, the explanations provided are not necessary minimal \cite{ignatiev2020towards}. Logic based rigorous approaches, on the other hand, are based on formal method and thus are provably accurate. Two main methodologies are adopted in logic based rigorous XAI/IAI approaches, one is knowledge compilation, the other is abductive reasoning \cite{ignatiev2020towards}. Adnan Darwiche and his team use knowledge compilation to compile binary models into Boolean circuits, and seek rigorous, logical complete, prime implicant explanations \cite{darwiche2020reasons}.

As model complexity is a concept often mentioned in the field of artificial intelligence, which emphasizes the complexity of the model in structure, thus heuristic methods can be classified into two categories by limiting the complexity of the model: one is ante-hoc, and the other is post-hoc. Ante-hoc is mainly for the model with lower complexity, and post-hoc is mainly for the model with higher complexity \cite{arrieta2020explainable,lipton2018mythos}. Examples of ``interpretable'' methods are tree-based models, linear regression, logistic regression, and Naive Bayes Model. Although the tree-based models are believed to be ``interpretable'' for they can provide the decision paths (or rules) that leads to the decisions, many researchers point out that the explanations are ``shallow'' and contain potentially too many redundant features and rules, and thus actually unable to provide irredundant sufficient reasons, which also known as \emph{prime implicant explanations}, or \emph{minimal sufficient reasons} \cite{DBLP2021}.

Heuristic interpretation methods mainly include Partial Dependence Plot (PDP) \cite{zhao2021causal}, Accumulated Local Effect Plot (ALEP) \cite{apley2020visualizing}, feature interaction, approximation model, local approximation model, shapely value, SHAP \cite{lundberg2017unified}. PDP and ALEP both describe how features affect the prediction of machine learning models, and can show whether the relationship between target and feature is linear, monotonic, or more complex. Among them, ALEP has faster calculation speed than PDP, and calculation deviation is smaller. The global proxy model is an interpretable model, which can approximate the prediction of the black box model after training. Black box models can be explained by explaining the proxy model. The purpose of the interpretable global proxy model is to approximate the prediction of the underlying model as accurate as possible, and it can be interpreted at the same time. The concept of proxy model can be found under different names: Approximation Model, Meta Model, Response Surface Model, Emulator, etc. The local proxy model is itself an interpretable model, used to explain the prediction of a single instance of the black box machine learning model. Local Interpretable Model-agnostic Explanations (LIME) \cite{ribeiro2016should}, Local Explanation Method-using Nonlinear Approximation (LEMNA) \cite{guo2018lemna} and other variations have been proposed as local approximation methods. By capturing the local features of the model and achieving good results in the interpretation of text and images, local proxy model methods have achieved good results in the interpretation of text and images by capturing the local features of the model. These agent models have the benefit of yielding simple explanations, but they focus on training a local agent model to explain a single prediction, and could not train a global agent model. The SHAP \cite{lundberg2017unified} based on the Shapley value determines its importance by calculating the individual's contribution, and it was tried by the Bank of England to explain the mortgage default model \cite{lundberg2017unified}. Google also combines Tensorflow with SHAP to further improve interpretability \cite{lundberg2017unified}. Rule argumentation based methods use user defined rules as an argument for reasoning system and express the explanation to the decision made. 

A brief summarization of XAI methodologies are presented in Table \ref{table_sum}, most of the current XAI approaches are based on heuristic methods, although they work well on the selected datasets, they cannot guarrenteen their performance on the other datasets.

\subsection{Explainable Artificial Intelligence in intrusion detection}

Various approaches trying to explain machine learning models for cybersecurity have been proposed \cite{Vigan2020ExplainableS, vadillo2021and, muna2021demystifying,8553598,mahbooba2021explainable,marino2018adversarial,grosse2017statistical}. Luca Vigano and his group proposed a new paradigm in security research called Explainable Security (XSec) in \cite{Vigan2020ExplainableS}. They propose the ``Six Ws''  of XSec (Who? What? Where? When? Why? and How?) as the standard perspectives of XAI in cybersecurity domain.

Marco Melis trys to explain malicious black-box android malware detections on any-type of models \cite{8553598}. This work leverages a gradient-based approach to identify the most influential local features. It also enables use of nonlinear models to potentially increase accuracy without sacrificing interpretability of decisions on the DREBIN Dataset. Drebin as such explains its decisions by reporting, for any given application, the most influential features, i.e., those features present in a given application and assigned the highest absolute weights by the classifier. 

\cite{marino2018adversarial} tries to use adversarial machine learning to find the minimum modifications (of the input features) required to correctly classify a given set of misclassified samples, to be specific, it tries to find an adversarial sample that is classified as positive with the minimum distance between the real sample and the modified sample.

Other works on explainable android malware detection do not specifically use explainable machine learning models but do make use of other feature analyses to reduce uncertainty of information. For example, \cite{grosse2017statistical} uses static analysis and probability statistics-based feature extraction analysis to detect and analyze malicious Android apps.

\cite{mahbooba2021explainable} try to interpret the rules for malicious node identification by directly use paths in decision tree model trained by KDD dataset. However, many researchers believe the direct interpretation provided by decision tree model is ``shallow'' and redundancy, and thus is not necessarily the minimal prime implicant interpretation, in other words, it is not the radical reason. 

A brief summary of current XAI applications on cybersecurity is presented in Table \ref{table_cyber}, with details of the methodology, datasets used, and target models. In the author's humble knowledge, almost all the state-of-the-art methodologies in intrusion detection are heuristic or simply direct ``shallow'' interpretations provided by tree-based models, both are neither accurate nor sufficient, thus cannot be considered really ``interpretable'' \cite{ignatiev2020towards,paredes2021importance}. Logic based rigorous approaches, on the other hand, are based on formal method and thus are provably accurate \cite{darwiche2020reasons}.

\subsection{DDoS Attacks}
DDoS is an attack that the attacker seeks to exhaust network resource by disrupting services of multiple hosts in the network. It is usually done by flooding the target hosts with superfluous requests attempting to overload the network. 

DDoS attack traffic flows show different patterns according to the tools that generate them. There are many tools available for DDoS attack generation, the top ones are:

\textbf{Low Orbit Ion Cannon (LOIC):} designed to flood target systems with junk TCP, UDP and HTTP GET requests. However, a single LOIC user is unable to generate enough requests to significantly impact a target. For an attack to succeed, thousands of users must coordinate and simultaneously direct traffic to the same network.

\textbf{High Orbit Ion Cannon (HOIC):} Designed to improve several LOIC flaws, HOIC is able to attack as many as 256 URLs at the same time. Unlike LOIC, which is able to launch TCP, UDP and HTTP GET floods, HOIC conducts attacks based solely on HTTP GET and POST requests. 

\textbf{SolarWinds\footnote{\url{https://www.solarwinds.com/security-event-manager/use-cases/ddos-attack?CMP=BIZ-RVW-SWTH-DDoSAttackTools-SEM-UC-Q120}}:} provides a security event manager that is effective mitigation and prevention software to stop the DDoS Attack. It will monitor the event logs from a wide range of sources for detecting and preventing DDoS activities. The security event manager will identify interactions with potential command and control servers by taking advantage of community-sourced lists of known bad actors. For this, it consolidates, normalizes, and reviews logs from various sources like IDS/IPs, firewalls, servers, etc.

\textbf{HTTP Unbearable Load King (HULK)\footnote{\url{https://packetstormsecurity.com/files/112856/HULK-Http-Unbearable-Load-King.html}}:} It is a DoS attack tool for the web server. It is created for research purposes. It can bypass the cache engine, generate unique and obscure traffic, but it may fail in hiding the identity. Traffic coming through HULK can be blocked.

\textbf{Tor's Hammer\footnote{\url{https://sourceforge.net/projects/torshammer/}}:} It is created for testing purposes for slow post attack.

\textbf{Slowloris\footnote{\url{https://github.com/gkbrk/slowloris}}:} Slowloris tool is used to make a DDoS attack. It is used to make the server down. It sends authorized HTTP traffic to the server while doesn?t affect other services and ports on the target network. This attack tries to keep the maximum connection engaged with those that are open by sending a partial request. It tries to hold the connections as long as possible. As the server keeps the false connection open, this will overflow the connection pool and will deny the request to the true connections. However, as it makes the attack at a slow rate, traffic can be easily detected and blocked.

\textbf{XOIC\footnote{\url{https://sourceforge.net/directory/?q=xoic}}:} a DDoS attacking tool that can fire attack on small websites. It is easy to use, but also easy to detect and block.

\textbf{DDoS Simulator (DDOSIM)\footnote{\url{https://sourceforge.net/projects/ddosim/}}:} works on Linux system, it is designed for simulating the real DDoS attack. It can attack on the website as well as on the network using valid or invalid requests. 

\textbf{R-U-Dead-Yet (RUDY)\footnote{\url{https://sourceforge.net/projects/r-u-dead-yet/}}:} makes the attack using a long form field submission through POST method. As it works at a very slow rate, it can be easily detected and blocked.

\emph{DDoS HOIC} attack, \emph{DDoS LOIC UDP} attack, and \emph{DDoS LOIC HTTP} attack datasets are collected and investigated in this paper. 

\section{Rigorous XAI Driven Intrusion Detection System}\label{sec_rin}

\begin{table*}[!htbp]
\scriptsize
\centering
\caption{Summary of XAI Methodologies in Intrusion Detection.}
\label{table_cyber}
{
\begin{tabular}{|c|c|c|c|c|}
\hline
Name & Description & Target Models  &Dataset& Ref. \\ \hline
\tabincell{c}{Marino \\ et.al.'s work} & \tabincell{c}{use adversarial machine learning to find the minimum \\modifications  (of the input features) required to correctly \\classify a given set of misclassified samples, \\by finding an adversarial sample that is classified \\ as positive while minimizing the distance \\between the real sample and the modified sample.} & DNN & NSL-KDD & \cite{marino2018adversarial} \\  \hline

\tabincell{c}{Marco Melis\\ et.al.'s work} &  \tabincell{c}{leveraging a gradient-based approach to\\ identify the most influential local features, \\apply on any black-box machine-learning model} & Any & DREBIN & \cite{8553598} \\ \hline

DeNNeS  & \tabincell{c}{An embedded, deep learning-based cybersecurity\\ expert system extracting refined rules\\ from a trained multilayer DNN, could either\\ function on a complete or incomplete dataset.} & DNN & \tabincell{c}{UCI's phishing websites dataset \\ Android malware dataset}  & \cite{mahdavifar2020dennes} \\ \hline
 
\tabincell{c}{J. N. Paredes\\ et.al.'s work} &\tabincell{c}{An vision of combining knowledge reasoning-driven \\ and data-driven approaches for XAI in cybersecurity}& N/A&\tabincell{c}{National Vulnerability Database \\ MITRE CVE \\ MITRE CWE \\ MITRE ATT\&CK} & \cite{paredes2021importance} \\ \hline

LEMNA & \tabincell{c}{LEMNA generates a small set of interpretable features to \\ explain the model by approximating a local area \\ of the complex deep learning decision boundary.} & DNN & \tabincell{c}{Binary dataset \\generated in \\BYTEWEIGHT\cite{bao2014byteweight}} & \cite{guo2018lemna} \\  \hline

\end{tabular}
}
\end{table*}

\subsection{Rigorous XAI}\label{sec_xai}

While heuristic XAI methods compute approximations of real explanations, rigorous explanations are guaranteed to be accurate and sufficient. Rigorous explanation methods compile machine learning models into Boolean circuits that can make the same decisions with the models. 

In rigorous XAI theory \cite{darwiche2020reasons,ignatiev2019abduction,ignatiev2020towards}, a classifier is a \emph{Boolean function} which can be represented by a propositional formula $\Delta$. An \emph{implicant} $\tau$ of a propositional formula $\Delta$ is a term that satisfies $\Delta$, namely $\tau \models \Delta$. A \emph{prime implicant} is an implicant that is not subsumed by any other implicants, that is there is no implicate $\tau'$ that contains a strict subset of the literals of $\tau$. \emph{Prime implicant} have been used to give rigorous explanations in XAI. Explanations given using \emph{prime implicant} are also called \emph{sufficient reasons}, which are defined formally below by \cite{darwiche2020reasons}.

\textbf{Definition 1 (Sufficient Reason \cite{darwiche2020reasons})} \emph{A sufficient reason for decision $\Delta_{\alpha}$ is a property of instance $\alpha$ that is also a prime implicant of $\Delta_{\alpha}$ ($\Delta_{\alpha}$ is $\Delta$ if the decision is positive and $\urcorner\Delta$ otherwise).}

A \emph{sufficient reason} (or \emph{prime implicant explanation}) is also the \emph{minimum explanation}. The major difference between \emph{sufficient reason} and \emph{prime implicant} is that \emph{sufficient reason} disclose the reasons of a certain instance while \emph{prime implicant} illustrate the essential characteristics of the model \cite{darwiche2020reasons}. \emph{Sufficient reason} explains the root cause of the decision for an instance, in terms of the \emph{prime implicants} involved. The decision will stay unchanged no matter how the other characters change, and none of its strict subsets can justify the decision. Please be noted that a decision may have multiple sufficient reasons, sometimes many \cite{darwiche2020reasons}.

The Quine--McCluskey algorithm (QMC) (also known as the \emph{method of prime implicants} or \emph{tabulation method}) is used in this paper to get the prime implicants from a Boolean expression. It is a classic boolean expression minimization method developed by Willard V. Quine in 1952 and extended by Edward J. McCluskey in 1956 \cite{quine1952problem}. Many of the state-of-the-art methods computing prime implicant from a boolean expression are variations of the Quine--McCluskey algorithm.

\subsection{Architecture of \sys}
\begin{figure*}[thpb]
      \centering
      \includegraphics[width=6in]{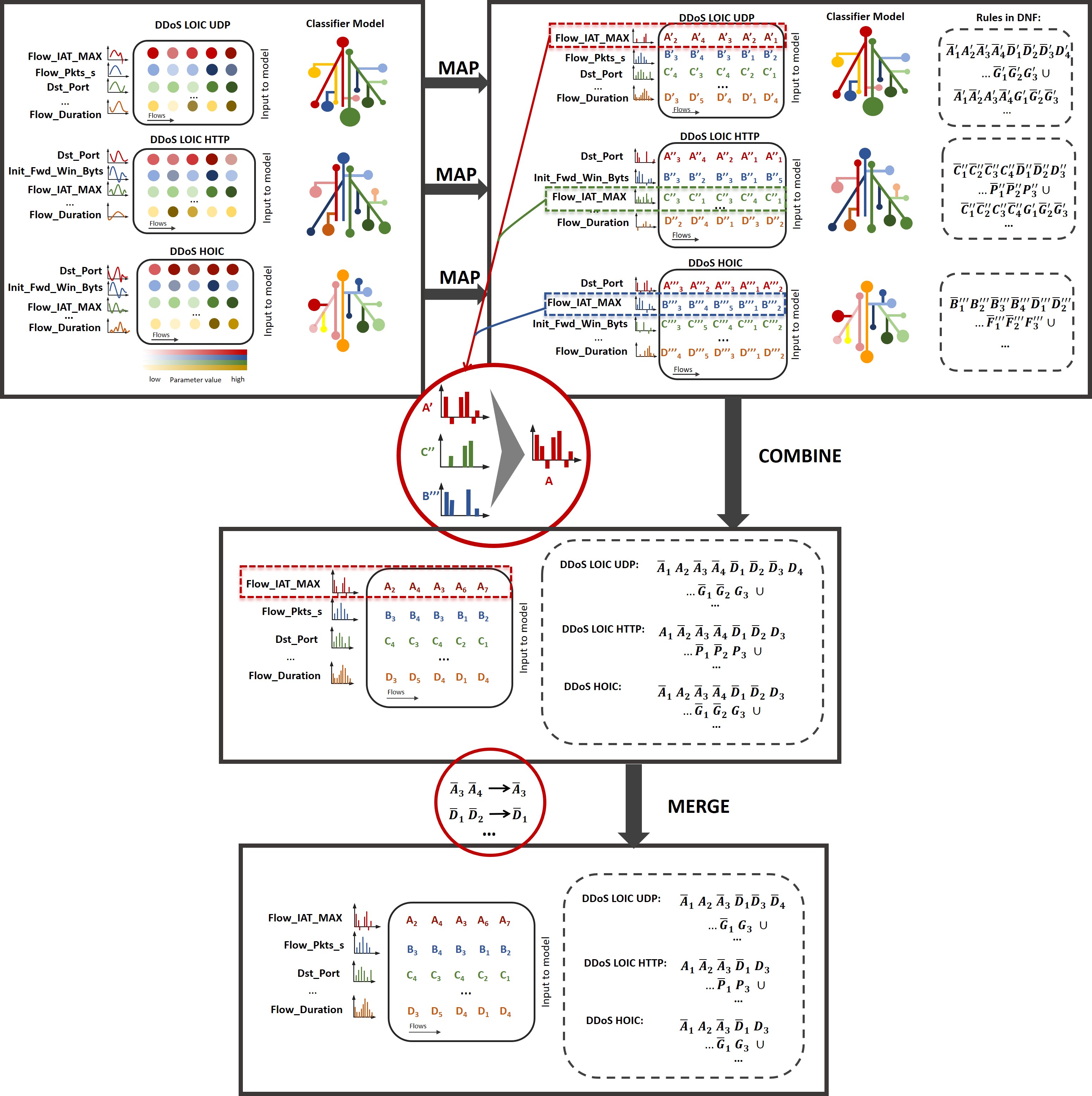} 
      \caption{The proposed architecture of \sys.}
      \label{fig_architecture}
\end{figure*}

The architecture of the proposed rigorous XAI driven intrusion detection system \sys~is shown in Fig. \ref{fig_architecture}. 

As discussed in our previous work \cite{zhou2019evaluation}, the flow-based statistical data generated from CICFlowMeter are used instead of direct packet header information. Features generated from CICFlowMeter, which have continuous values, are mapped into discrete variables as discussed in Section \ref{subsec_discretize}. Then boolean expression of the machine learning model will be generated and further simplified into prime implicants. Depend on the specific boolean expression, the simplification process, which is a SAT quesion, may have NP-hard complexity. The \emph{prime implicant} generated are the \emph{sufficient reasons} learned by the machine learning model from the flow-based statistical traffic data. They should be interrogated by human expert, judging with the experts' knowledge and experience. The audited rules should be taken into consideration when designing new informed machine learning models, to achieve more accurate intrusion detection.

\subsection{Formal Description of Map and Merge (M\&M)}\label{subsec_discretize}

\begin{figure}[htbp] 
	\subfigure[A Example of Decision Tree]
	{
		\begin{minipage}[t]{1.3in}\label{fig_demotree_a}
			\centering
			\includegraphics[width=1.3in]{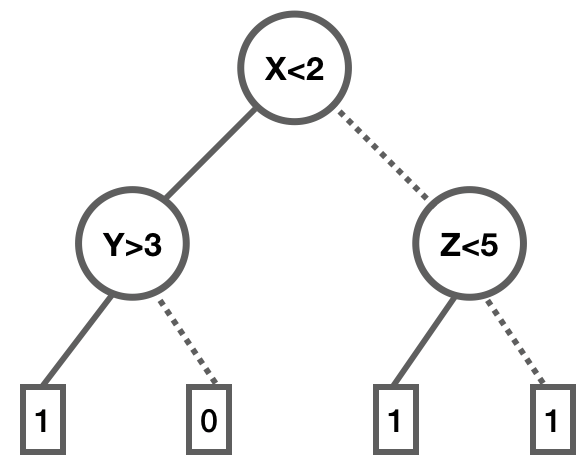} 
		\end{minipage}
	} 
	\quad 
	\subfigure[Discretized Features]{ 
		\begin{minipage}[t]{1.3in}\label{fig_demotree_b}
			\centering
			\includegraphics[width=1.3in]{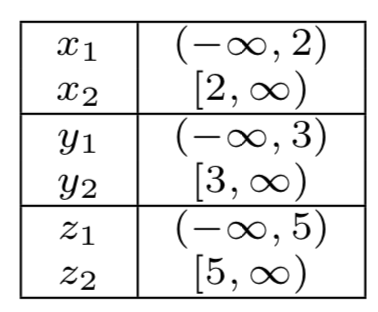} 
		\end{minipage}
	} 
	\caption{A simple decision tree with continuous values.} 
	\label{fig_demotree}
\end{figure}

Rigorous logical reasoning methods work directly on boolean expressions. Machine learning models with boolean features can be immediately represented as boolean circuits \cite{darwiche2020reasons}. Machine learning models with discrete features can be transformed into boolean expressions by representing the fact that ``a feature equals to a certain value'' with an atom variable \cite{darwiche2020reasons}. However, most of the classifiers used in intrusion detection system have continuous features. Darwiche et.al. proposed a mapping method to map continuous features into discrete ones \cite{darwiche2020reasons}, as presented below in Fig. \ref{fig_demotree}. Based on that mapping method, we proposed a \sys~algorithm for discretized feature for meta-learning models, by add combine and simplify process, hence map, combine, and merge (M\&M) method. The detail of \sys~discretize method is presented below. 

\subsubsection{Map}
Before describe the process of \sys~in detail, some formal definitions are given below.

\textbf{Definition \emph{Decision Path:}} A decision path is a conjunction of conditions from root node to the leaf. For a decision tree model with $n$ leafs, there are $n$ decision paths. 

\textbf{Definition \emph{Rule:}} A rule $\mathfrak{r}$ in a decision tree model is defined in the form of ``if $\mathcal{F}_{1} > \mathcal{F}_{1_{down}}$ and $\mathcal{F}_{1} < \mathcal{F}_{1_{up}}$ and \dots and $\mathcal{F}_{n} > \mathcal{F}_{n_{down}}$ and $\mathcal{F}_{n} < \mathcal{F}_{n_{up}}$ then $Label_{i}$'', where $\mathcal{F}_i$ is the $i th$ feature on the decision path and $\mathcal{F}_{i_{up}}$ is the upper bound of the feature value in the decision path, and $\mathcal{F}_{i_{down}}$ is the lower bound of the feature value.

Based on the definitions, the decision tree model can be defined as a set of rules $\mathfrak{R} = \{\mathfrak{r}_i\}$, $0 \leqslant i \leqslant n$, where $n$ is the number of decision paths in the model.

\begin{algorithm}[H]
\caption{Rule of Map}\label{alg_map}
\begin{algorithmic}
\STATE LET Feature Space: $\mathfrak{F}$ $\longleftarrow \varnothing $ \\
\STATE LET rule Set: $\mathfrak{R}$ $\longleftarrow$ $\{\mathfrak{r}_i \}$, $0 \leqslant i \leqslant n$ \\
FOR $\mathfrak{r}$ IN $\mathfrak{R}$: \\
\quad FOR features $\mathfrak{f}_i$ IN rule $\mathfrak{r}$: \\
\quad \quad $\mathfrak{F} \gets \mathfrak{f}_i$   \\ 
\quad \quad $\mathfrak{f}_i.values$  $\longleftarrow \varnothing$  \\
\quad \quad FOR every time $\mathfrak{f}_i$ appear IN rule $\mathfrak{r}$: \\
\quad \quad \quad $\mathfrak{f}_i.values$ $\gets$  $\mathfrak{f}_i.value$  \\ 
SORT $\mathfrak{f}_i.values$ \\
FOR intervals $(\mathfrak{f}_{i_{down}}, \mathfrak{f}_{i_{up}}]$ in $\mathfrak{f}_i.values$: \\
\quad ASSIGN discrete features $\mathfrak{Df}_i = \mathfrak{f}_i \in (\mathfrak{f}_{i_{down}}, \mathfrak{f}_{i_{up}}]$ \\
RETURN Discrete Features $\mathfrak{Df} = \{\mathfrak{Df}_i\}$

\end{algorithmic}
\end{algorithm}

Take the decision tree in Fig. \ref{fig_demotree_a} for example, the solid lines represent if the node is true, while the dashed lines represent false. Based on each decision node, features with continuous values are discretized into several variables, each represent an interval divided by decision nodes. As shown in Fig. \ref{fig_demotree_b}, feature $X$ in Fig. \ref{fig_demotree_a} are discretized into $x_1$, and $x_2$, representing the intervals $(-\infty, 2)$ and $[2, \infty)$ respectively, thus we have 

\begin{align}
{x_1}\vee{x_2} \models U~ \&\& ~{x_1}\wedge{x_2} \models \oslash \label{eq_1a}\\
{y_1}\vee{y_2} \models U~ \&\& ~ {y_1}\wedge{y_2} \models \oslash  \label{eq_1b}\\
{z_1}\vee{z_2} \models U~ \&\& ~{z_1}\wedge{z_2} \models \oslash \label{eq_1c}
\end{align}

Thus, the decision rule of the decision tree in Fig. \ref{fig_demotree_a} can be represented by boolean expression 

\begin{equation}\label{eq_demo}
\Delta = ({x_1}\wedge{y_1})\vee({x_2}\wedge{z_1})\vee({x_2}\wedge{z_2})
\end{equation}

According to Eq. \ref{eq_1c}, we have 

$$ {z_1} = \bar{z_2}$$

Together with De Morgan's law, Eq. \ref{eq_demo} can be further simplified to 

\begin{equation}\label{eq_demo2}
\Delta = ({x_1}\wedge{y_1})\vee{x_2}
\end{equation}

In which ${x_1}\wedge{y_1}$ and $x_2$ are \emph{prime implicants} of the decision tree in Fig. \ref{fig_demotree_a}, and thus the rigorous explanation of the decision tree are \emph{``So long as ${x_1}\wedge{y_1}$ or $x_2$, the decision will be 1''}. According to the discretization rule, the rigorous explanation can further be \emph{``So long as $X<2$ and $Y<3$, or $X\ge 5$ in the instance, the decision is guarranteed to be 1.''}

\subsubsection{Combine}

As intrusion datasets are often collected separately and distributively over the networking system, meta-learning is often required for each network element to build their own model and collect their own data and train their classifier as accurate as possible, and then all the models can be combined into a unified model ready for interpretation. The meta-learning model combining algorithm is shown in Algorithm \ref{alg_combine} in detail.

\begin{algorithm}[H]
\caption{Rule of Combine}\label{alg_combine}
\begin{algorithmic}
\STATE LET $\mathfrak{T}_i$ $\longleftarrow$ One model to combine \\
LET $\mathfrak{T}_j \longleftarrow$ Another model to combine \\
LET $\mathfrak{F}_i \longleftarrow$ features in $\mathfrak{T}_i$: \\
LET $\mathfrak{F}_j \longleftarrow$ features in $\mathfrak{T}_j$: \\ 
LET $\mathfrak{F}_{combine} \longleftarrow \varnothing$ \\
FOR feature $\mathfrak{f}_i$ in $\mathfrak{F}_i$: \\
\quad IF $\mathfrak{f}_i$ NOT IN $\mathfrak{F}_j$: \\
\quad \quad $\mathfrak{F}_{combine} \longleftarrow \mathfrak{f}_i$ \\
\quad ELSE: \\
\quad \quad $\mathfrak{F}_{combine} \longleftarrow$ merged sets $\mathfrak{f}_i \cup \mathfrak{f}_j$ 

RETURN $\mathfrak{F}_{combine}$
\end{algorithmic}
\end{algorithm}

\subsubsection{Merge}
The discrete features get from the map process may (and often) contain (potentially a large number of) redundancy. As the number of boolean expression is $$N = 2^n$$ where $n$ is the number of discrete features, directly transform the discrete features into boolean circuits may experience a huge waste of computing and storage expense, due to \emph{Combinatorial Explosion}. Thus the merge process proposed in Algorithm \ref{alg_merge} is used in \sys. 

\begin{algorithm}[H]
\caption{Rule of Merge}\label{alg_merge}
\begin{algorithmic}
\STATE For $x_i = (v_1, v_2] $ and $x_{i+1} = (v_2,  v_3]$, $(v_1 \leq v_2 \leq v_3)$
\STATE AND $s_i = \{set~of~features~in~rule~r_i\}$
\STATE IF  $\neg \exists ((x_i \in s_i) \wedge (x_{i+1} \notin s_i)) \lor ((x_i \notin s_i) \wedge (x_{i+1} \in s_i)) $ 
\STATE THEN DELETE $x_{i+1}$ AND $x_i = x_i \cup x_{i+1} = (v_1, v_3] $
\end{algorithmic}
\end{algorithm}

For example, let $\Delta = ((x_1 \vee x_2) \wedge y_1) \vee ((x_1 \vee x_2) \wedge z_2)$ be the boolean expression of a model after map process, according to Algorithm \ref{alg_merge}, $(x_1 \vee x_2)$ fits the requirement of merge rule, and thus the model can be simplified into $\Delta = (x_{1new} \wedge y_1) \vee (x_{1new} \wedge z_2)$, where $x_{1new} = x_{1old} \vee x_{2old}$.

\section{Evaluation of \sys}\label{sec_res}
Evaluation experiments are carried out and presented in this section. 

\subsection{Model Selection}

In our previous work, we have evaluated 8 kinds of common machine learning models on eleven different kinds of real-life intrusion traffic data \cite{zhou2019evaluation}. From the evaluation result in \cite{zhou2019evaluation}, it is evident that decision tree has the best performance in both accuracy and time expense when detecting known intrusion. Thus we use decision tree model trained in the previous experiment as the target model, and compute the rigorous explanation of the model.

\begin{figure}[thpb]
      \centering
      \includegraphics[width=2.5in]{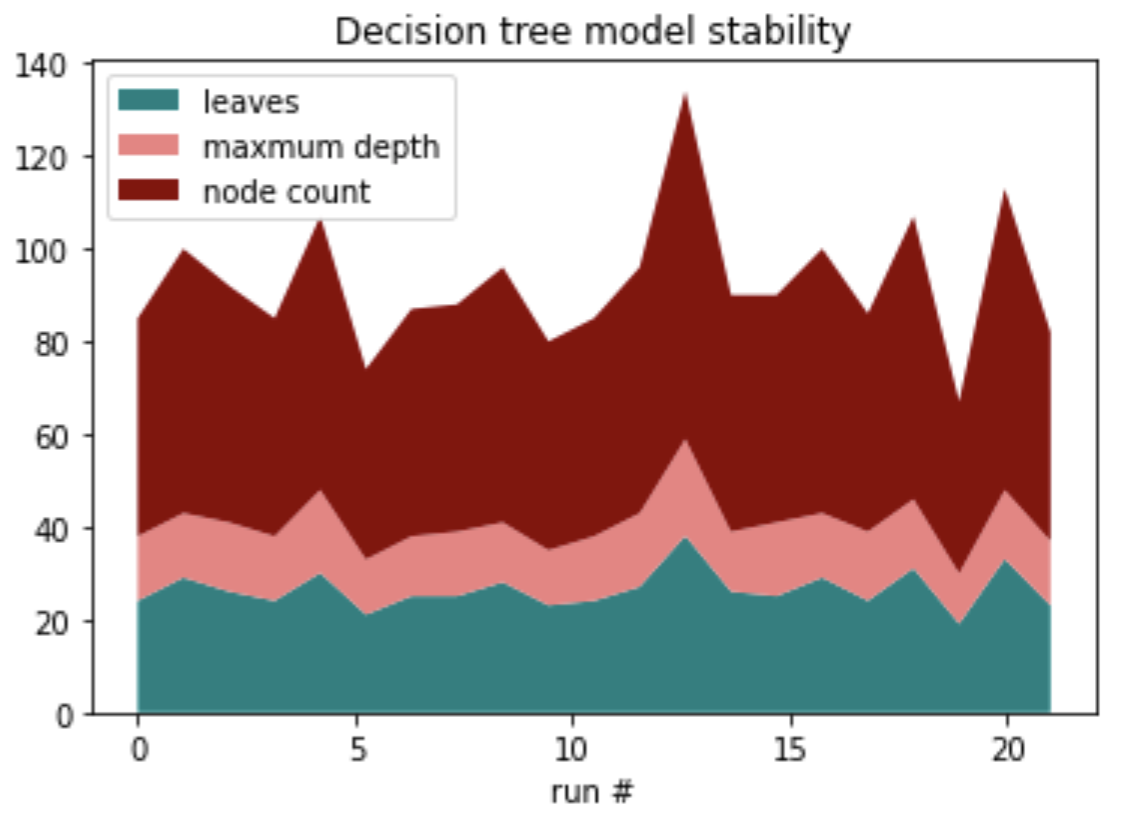} 
      \caption{The stability of decision tree model.}
      \label{fig_stable}
\end{figure}

The decision tree model varies during the fitting process. A model is considered to be stable when the features of model (e.g., number of leaves in the tree, maximum depth, and node count) are tending towards stability. For example, if the number of leaves (or maximum depth, node count, etc.) in the tree keeps growing with the fitting process, it implies that the model is still learning new rules from the training data, and when the features fluctuates around a certain value, it implies the model has learnt all the rules from training dataset and is considered to be stable. The training process is carried out and the model features together with the detection results are recorded and represented in Fig. \ref{fig_stable}. The features of the decision tree models used are \emph{number of leaves in the tree, maximum depth,} and \emph{node count}. The indicators for detection accuracy are \emph{area under the curve (AUC), precision of benign traffic, precision of evil traffic}. 

\begin{figure*}[] 
	\subfigure[Features correlation of DDos LOIC UDP \& HTTP]
	{
		\begin{minipage}[t]{3in}\label{fig_demotree_a}
			\centering
			\includegraphics[width=3in]{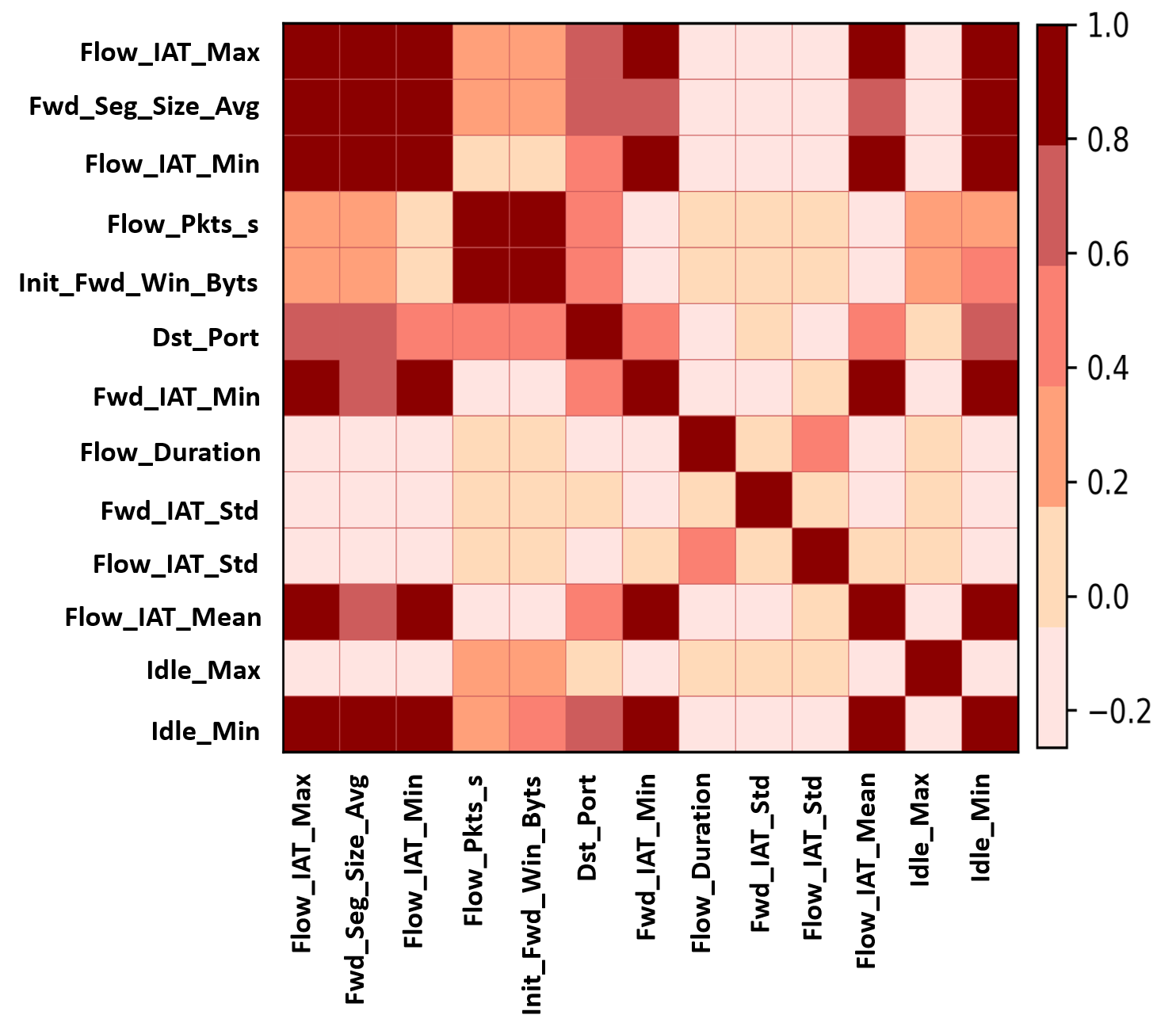} 
		\end{minipage}
	} 
	\quad 
	\subfigure[Feature correlation of DDoS HOIC]{ 
		\begin{minipage}[t]{3in}\label{fig_demotree_b}
			\centering
			\includegraphics[width=3in]{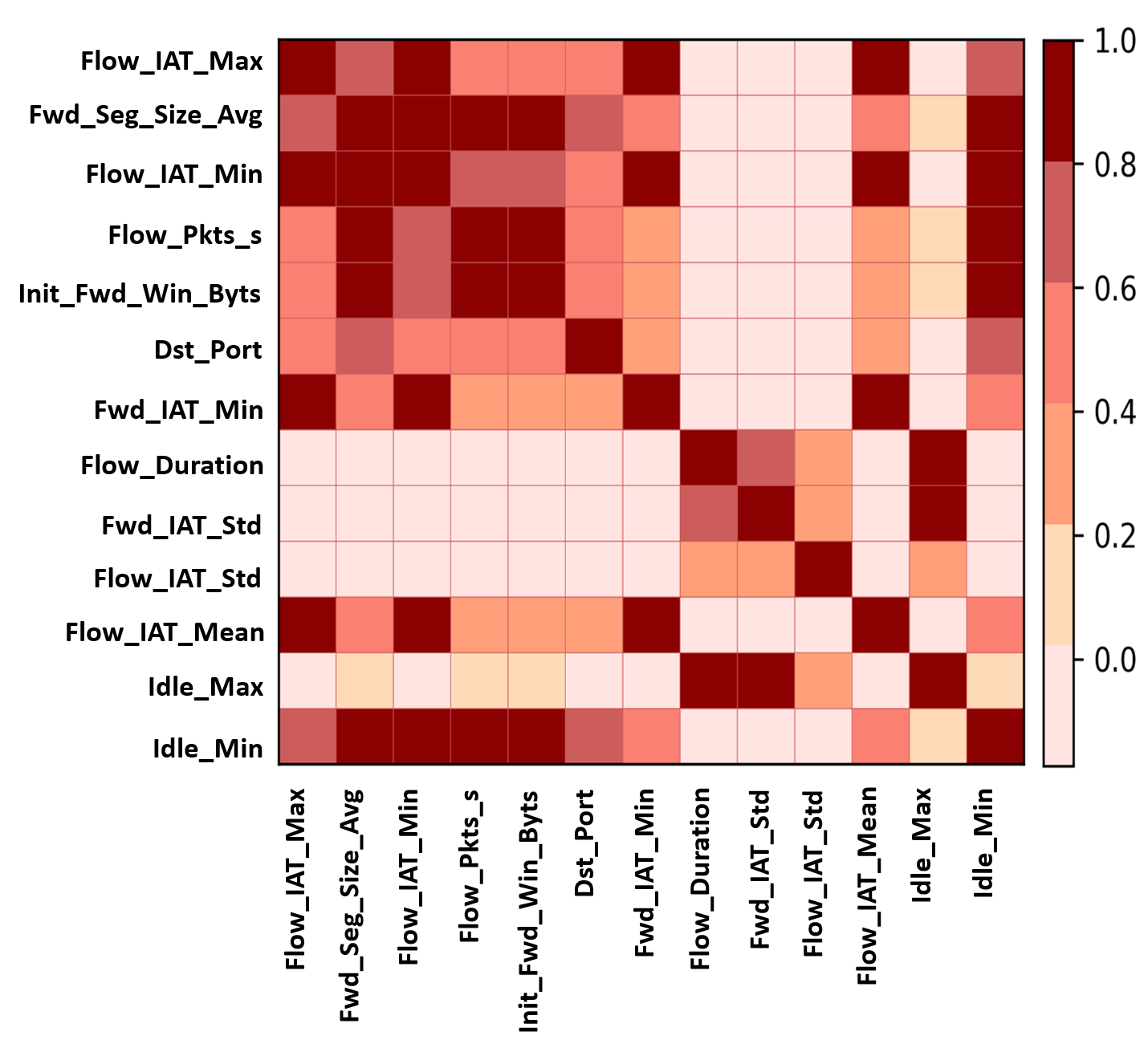} 
		\end{minipage}
	} 
	\caption{A simple decision tree with continuous values.} 
	\label{fig_corr}
\end{figure*} 

From Fig. \ref{fig_stable} it is evident that the \emph{number of leaves, maximum depth}, and \emph{node count} of decision tree models trained per round fluctuate within a narrow range, do not show any pronounce trend (of increase or decline). It is reasonable to believe that the model is stable and all that could be learn from the training dataset have been learn. A modest model (by modest I mean in terms of \emph{number of leaves, maximum depth}, and \emph{node count}) is selected as the target model for rigorous explanation computing. The correlation between features of the selected models are shown in Fig. \ref{fig_corr}.

\subsection{Feature Discretization with M\&M}
M\&M discretization method in Section \ref{subsec_discretize} is used to transform continuous features into discretized one. The Features before and after discretization are presented in Table \ref{table_treefeatures}. After discretization, 16 continuous features are transformed to 47 discrete variables. Thus, based on their feature, each instance can be mapped into a 47 bits binary expression 
$${a_1}{a_2}{b_1}{b_2}{c_1}{c_2}{d_1}{d_2}{d_3}{d_4}{d_5}{e_1}{e_2}{f_1}{f_2}{f_3}{f_4}{f_5}{f_6}{f_7}{f_8}{g_1}{g_2}{g_3}$$$${g_4}{g_5}{g_6}{h_1}{h_2}{h_3}{i_1}{i_2}{i_3}{j_1}{j_2}{k_1}{k_2}{l_1}{l_2}{m_1}{m_2}{n_1}{n_2}{o_1}{o_2}{p_1}{p_2}$$

\begin{table}[ht]
\scriptsize
\centering
\caption{Discretized Feature after \sys~process.}
\label{table_treefeatures}
\resizebox{.5\textwidth}{70mm}
{
\begin{tabular}{|c|c|c|}
\hline
\textbf{Continuous} & \textbf{After Map} & \textbf{After M\&M} \\ \hline
\multirow{2}*{A: Subflow\_Fwd\_Byts} & $a_1$: $(-\infty, 21]$ & $a_1$: $(-\infty, 21]$ \\ \cline{2-3}
& $a_2$: $(21, +\infty)$ & $a_2$: $(21, +\infty)$ \\ \hline

\multirow{2}*{B: Flow\_IAT\_Max} & $b_1$: $(-\infty, 342916]$ & $b_1$: $(-\infty, 342916]$ \\ \cline{2-3}
& $b_2$: $(342916, +\infty)$ & $b_2$: $(342916, +\infty)$ \\ \hline

\multirow{2}*{C: Fwd\_Seg\_Size\_Avg} & $c_1$: $(-\infty, 5.7]$ & $c_1$: $(-\infty, 5.7]$ \\ \cline{2-3}
& $c_2$: $(5.7, +\infty)$ & $c_2$: $(5.7, +\infty)$ \\ \hline

\multirow{7}*{D: Flow\_IAT\_Min} & $d_1$: $(-\infty, 1.5]$ & \multirow{3}*{$d_{1-3}$: $(-\infty, 5440.5]$} \\ \cline{2-2}
& $d_2$: $(1.5, 4997.5]$ &  \\ \cline{2-2}
& $d_3$: $(4997.5, 5440.5]$ &   \\ \cline{2-3}
& $d_4$: $(4997.5, 5440.5]$ & $d_4$: $(4997.5, 5440.5]$  \\ \cline{2-3}
& $d_5$: $(5836, 11980.5]$ & $d_5$: $(5836, 11980.5]$ \\ \cline{2-3}
& $d_6$: $(11980.5, 56360494]$ & $d_6$: $(11980.5, 56360494]$ \\ \cline{2-3}
& $d_7$: $(56360494, +\infty]$ & $d_7$: $(56360494, +\infty]$ \\ \hline

\multirow{2}*{E: Flow\_Pkts\_s} & $e_1$: $(-\infty, 367.35]$ & $e_1$: $(-\infty, 367.35]$ \\ \cline{2-3}
& $e_2$: $(367.35, +\infty)$ & $e_2$: $(367.35, +\infty)$ \\ \hline

\multirow{9}*{F: Init\_Fwd\_Win\_Byts} & $f_1$: $(-\infty, 252.5]$ & \multirow{2}*{$f_{1-2}$: $(-\infty, 1794]$} \\ \cline{2-2}
& $f_2$: $(252.5, 1794]$ & \\ \cline{2-3}
& $f_3$: $(1794, 1948]$ & $f_3$: $(1794, 1948]$ \\ \cline{2-3}
& $f_4$: $(1948, 1999.5]$ & $f_4$: $(1948, 1999.5]$ \\ \cline{2-3}
& $f_5$: $(1999.5, 2079.5]$ & $f_5$: $(1999.5, 2079.5]$ \\ \cline{2-3}
& $f_6$: $(2079.5, 2520.5]$ & $f_6$: $(2079.5, 2520.5]$ \\ \cline{2-3}
& $f_7$: $(2520.5, 5120.5]$ & $f_7$: $(2520.5, 5120.5]$ \\ \cline{2-3}
& $f_8$: $(5120.5, 30969]$ & $f_8$: $(5120.5, 30969]$ \\ \cline{2-3}
& $f_9$: $(30969, +\infty)$ & $f_9$: $(30969, +\infty)$ \\ \hline

\multirow{6}*{G: Dst\_Port} & $g_1$: $(-\infty, 51.5]$ & $g_{1}$: $(-\infty, 52.5]$ \\ \cline{2-3}
& $g_2$: $(51.5, 52.5]$ &  $g_2$: $(51.5, 52.5]$ \\ \cline{2-3}
& $g_3$: $(52.5, 80.5]$ & $g_3$: $(52.5, 80.5]$ \\ \cline{2-3}
& $g_4$: $(80.5, 261.5]$ & $g_4$: $(80.5, 261.5]$ \\ \cline{2-3}
& $g_5$: $(261.5, 25929]$ & $g_5$: $(261.5, 25929]$ \\ \cline{2-3}
& $g_6$: $(25929, +\infty)$ & $g_6$: $(25929, +\infty)$ \\ \hline

\multirow{3}*{H: Fwd\_IAT\_Min} & $h_1$: $(-\infty, 16.5]$ & $h_1$: $(-\infty, 16.5]$ \\ \cline{2-3}
& $h_2$: $(16.5, 92921]$ & $h_2$: $(16.5, 92921]$ \\ \cline{2-3}
& $h_3$: $(92921, +\infty)$ & $h_3$: $(92921, +\infty)$ \\ \hline

\multirow{3}*{I: Flow\_Duration} & $i_1$: $(-\infty, 52151692]$ & $i_1$: $(-\infty, 52151692]$ \\ \cline{2-3}
& $i_2$: $(52151692, 53010668]$ & $i_2$: $(52151692, 53010668]$ \\ \cline{2-3}
& $i_3$: $(53010668, +\infty)$ & $i_3$: $(53010668, +\infty)$ \\ \hline

\multirow{2}*{J: Fwd\_IAT\_Std} & $j_1$: $(-\infty, 16847.75]$ & $j_1$: $(-\infty, 16847.75]$ \\ \cline{2-3}
& $j_2$: $(16847.75, +\infty)$ & $j_2$: $(16847.75, +\infty)$ \\ \hline

\multirow{2}*{K: Flow\_IAT\_Std} & $k_1$: $(-\infty, 978470.19]$ & $k_1$: $(-\infty, 978470.19]$ \\ \cline{2-3}
& $k_2$: $(978470.19, +\infty)$ & $k_2$: $(978470.19, +\infty)$ \\ \hline
\multirow{2}*{L: Flow\_IAT\_Mean} & $l_1$: $(-\infty, 16000000]$ & $l_1$: $(-\infty, 16000000]$ \\ \cline{2-3}
& $l_2$: $(16000000, +\infty)$ & $l_2$: $(16000000, +\infty)$ \\ \hline

\multirow{2}*{M: Idle\_Max} & $m_1$: $(-\infty, 52194244]$ & $m_1$: $(-\infty, 52194244]$ \\ \cline{2-3}
& $m_2$: $(52194244, +\infty)$ & $m_2$: $(52194244, +\infty)$ \\ \hline

\multirow{3}*{N: Idle\_Min} & $n_1$: $(-\infty, 28200000]$ & $n_1$: $(-\infty, 28200000]$ \\ \cline{2-3}
& $n_2$: $(28200000, 53100000]$ & \multirow{2}*{$n_{2-3}$: $(28200000, +\infty)$} \\ \cline{2-2} 
& $n_3$: $(53100000, +\infty)$ & \\ \hline

\multirow{2}*{O: ACK\_Flag\_Cnt} & $o_1$: $(-\infty, 0.5]$ & $o_1$: $(-\infty, 0.5]$ \\ \cline{2-3}
& $o_2$: $(0.5, +\infty)$ & $o_2$: $(0.5, +\infty)$ \\ \hline

\multirow{2}*{P: Fwd\_Act\_Data\_Pkts} & $p_1$: $(-\infty, 201.5]$ & $p_1$: $(-\infty, 201.5]$ \\ \cline{2-3}
& $p_2$: $(201.5, +\infty)$ & $p_2$: $(201.5, +\infty)$  \\ \hline

\end{tabular}
}
\end{table}

\begin{algorithm}[h]
\caption{Continuous Features of a sample flow.}\label{alg:alg1}
\begin{algorithmic}
\STATE $Subflow\_Fwd\_Byts = 553$, 
\STATE $Flow\_IAT\_Max = 73403 $, 
\STATE $Fwd\_Seg\_Size\_Avg= 61.4 $, $Flow\_Pkts\_s = 113.2 $, 
\STATE $Init\_Fwd\_Win\_Byts = 8192 $, $Dst\_Port = 22 $, 
\STATE $ Fwd\_IAT\_Min = 14 $, $Flow\_Duration = 88751 $, 
\STATE $ Fwd\_IAT\_Std = 49295.7 $, $Flow\_IAT\_Std = 42321 $, 
\STATE $ Flow\_IAT\_Mean = 42321 $, $ Idle\_Max = 0.0 $, 
\STATE $Idle\_Min = 0 $, $Ack\_Flag\_Cnt = 0.0 $
\STATE $Fwd\_Act\_Data\_Pkts = 7$
\end{algorithmic}
\label{alg1}
\end{algorithm}

\begin{table*}[!htbp]
\scriptsize
\centering
\caption{Prime Implicants explanations for DDoS LOIC HTTP attack (1/2).}
\label{table_loichttp1}
{
\bgroup
\def\arraystretch{1.5}
\begin{tabular}{|c|c|c|}
\hline
\# &Minterm & Boolean Expression \\ \hline
$\tau_1$ & $a_1 \bar{a_2} b_1 \bar{b_2} c_1 \bar{c_2} \bar{d_1} d_2 \bar{d_3} \bar{d_4} \bar{d_5} \bar{e_1} e_2 $ & $1010100100001----------------------------------$  \\ \hline

$\tau_2$ & $a_1 \bar{a_2} b_1 \bar{b_2} c_1 \bar{c_2} \bar{d_1} \bar{d_2} \bar{d_3} \bar{d_4} {d_5} \bar{f_1} \bar{f_2} f_3 \bar{f_4} \bar{f_5} \bar{f_6} \bar{f_7} \bar{f_8} $ & $10101000100--00100000--------------------------$ \\ \hline
$\tau_3$ & $a_1 \bar{a_2} b_1 \bar{b_2} c_1 \bar{c_2} \bar{d_1} \bar{d_2} \bar{d_3} \bar{d_4} {d_5} \bar{f_1} \bar{f_2} \bar{f_3} {f_4} \bar{f_5} \bar{f_6} \bar{f_7} \bar{f_8} $ & $10101000100--00010000--------------------------$ \\ \hline
$\tau_4$ & $a_1 \bar{a_2} b_1 \bar{b_2} c_1 \bar{c_2} \bar{d_1} \bar{d_2} \bar{d_3} \bar{d_4} {d_5} \bar{f_1} \bar{f_2} \bar{f_3} \bar{f_4} {f_5} \bar{f_6} \bar{f_7} \bar{f_8} $ & $10101000100--00001000--------------------------$ \\ \hline
$\tau_5$ & $a_1 \bar{a_2} b_1 \bar{b_2} c_1 \bar{c_2} \bar{d_1} \bar{d_2} \bar{d_3} \bar{d_4} {d_5} \bar{f_1} \bar{f_2} \bar{f_3} \bar{f_4} \bar{f_5} {f_6} \bar{f_7} \bar{f_8} $ & $10101000100--00000100--------------------------$ \\ \hline

$\tau_6$ & $ a_1 \bar{a_2} {b_1} \bar{b_2} {c_1} \bar{c_2} \bar{d_1} \bar{d_2} \bar{d_3} {d_4} \bar{d_5} f_1 \bar{f_2} \bar{f_3} \bar{f_4} \bar{f_5} \bar{f_6} \bar{f_7} \bar{f_8} {g_1} \bar{g_2} \bar{g_3} \bar{g_4} \bar{g_5} \bar{g_6} \bar{h_1} \bar{h_2} h_3 $ & $10101000010--10000000100000001-----------------$ \\ \hline
$\tau_7$ & $ a_1 \bar{a_2} {b_1} \bar{b_2} {c_1} \bar{c_2} \bar{d_1} \bar{d_2} \bar{d_3} \bar{d_4} {d_5} f_1 \bar{f_2} \bar{f_3} \bar{f_4} \bar{f_5} \bar{f_6} \bar{f_7} \bar{f_8} {g_1} \bar{g_2} \bar{g_3} \bar{g_4} \bar{g_5} \bar{g_6} \bar{h_1} \bar{h_2} h_3 $ & $10101000001--10000000100000001-----------------$ \\ \hline
$\tau_8$ & $ a_1 \bar{a_2} {b_1} \bar{b_2} {c_1} \bar{c_2} \bar{d_1} \bar{d_2} \bar{d_3} {d_4} \bar{d_5} f_1 \bar{f_2} \bar{f_3} \bar{f_4} \bar{f_5} \bar{f_6} \bar{f_7} \bar{f_8} \bar{g_1} {g_2} \bar{g_3} \bar{g_4} \bar{g_5} \bar{g_6} \bar{h_1} \bar{h_2} h_3 $ & $10101000010--10000000010000001-----------------$ \\ \hline
$\tau_9$ & $ a_1 \bar{a_2} {b_1} \bar{b_2} {c_1} \bar{c_2} \bar{d_1} \bar{d_2} \bar{d_3} \bar{d_4} {d_5} f_1 \bar{f_2} \bar{f_3} \bar{f_4} \bar{f_5} \bar{f_6} \bar{f_7} \bar{f_8} \bar{g_1} {g_2} \bar{g_3} \bar{g_4} \bar{g_5} \bar{g_6} \bar{h_1} \bar{h_2} h_3 $ & $10101000001--10000000010000001-----------------$ \\ \hline
$\tau_{10}$ & $ a_1 \bar{a_2} {b_1} \bar{b_2} {c_1} \bar{c_2} \bar{d_1} \bar{d_2} \bar{d_3} \bar{d_4} {d_5} f_1 \bar{f_2} \bar{f_3} \bar{f_4} \bar{f_5} \bar{f_6} \bar{f_7} \bar{f_8} \bar{g_1} \bar{g_2} {g_3} \bar{g_4} \bar{g_5} \bar{g_6} \bar{h_1} \bar{h_2} h_3 $ & $10101000010--10000000001000001-----------------$ \\ \hline
$\tau_{11}$ & $ a_1 \bar{a_2} {b_1} \bar{b_2} {c_1} \bar{c_2} \bar{d_1} \bar{d_2} \bar{d_3} \bar{d_4} {d_5} f_1 \bar{f_2} \bar{f_3} \bar{f_4} \bar{f_5} \bar{f_6} \bar{f_7} \bar{f_8} \bar{g_1} \bar{g_2} {g_3} \bar{g_4} \bar{g_5} \bar{g_6} \bar{h_1} \bar{h_2} h_3 $ & $10101000001--10000000001000001-----------------$ \\ \hline
$\tau_{12}$ & $ a_1 \bar{a_2} {b_1} \bar{b_2} {c_1} \bar{c_2} \bar{d_1} \bar{d_2} \bar{d_3} {d_4} \bar{d_5} f_1 \bar{f_2} \bar{f_3} \bar{f_4} \bar{f_5} \bar{f_6} \bar{f_7} \bar{f_8} \bar{g_1} \bar{g_2} \bar{g_3} {g_4} \bar{g_5} \bar{g_6} \bar{h_1} \bar{h_2} h_3 $ & $10101000010--10000000000100001-----------------$ \\ \hline
$\tau_{13}$ & $ a_1 \bar{a_2} {b_1} \bar{b_2} {c_1} \bar{c_2} \bar{d_1} \bar{d_2} \bar{d_3} \bar{d_4} {d_5} f_1 \bar{f_2} \bar{f_3} \bar{f_4} \bar{f_5} \bar{f_6} \bar{f_7} \bar{f_8} \bar{g_1} \bar{g_2} \bar{g_3} {g_4} \bar{g_5} \bar{g_6} \bar{h_1} \bar{h_2} h_3 $ & $10101000001--10000000000100001-----------------$ \\ \hline

$\tau_{14}$ & $ a_1 \bar{a_2} {b_1} \bar{b_2} c_1 \bar{c_2} \bar{d_1} \bar{d_2} \bar{d_3} {d_4} \bar{d_5} \bar{f_1} {f_2} \bar{f_3} \bar{f_4} \bar{f_5} \bar{f_6} \bar{f_7} \bar{f_8} $ & $10101000010--01000000--------------------------$ \\ \hline
$\tau_{15}$ & $ a_1 \bar{a_2} {b_1} \bar{b_2} c_1 \bar{c_2} \bar{d_1} \bar{d_2} \bar{d_3} \bar{d_4} {d_5} \bar{f_1} {f_2} \bar{f_3} \bar{f_4} \bar{f_5} \bar{f_6} \bar{f_7} \bar{f_8} $ & $10101000001--01000000--------------------------$ \\ \hline
$\tau_{16}$ & $ a_1 \bar{a_2} {b_1} \bar{b_2} c_1 \bar{c_2} \bar{d_1} \bar{d_2} \bar{d_3} {d_4} \bar{d_5} \bar{f_1} \bar{f_2} {f_3} \bar{f_4} \bar{f_5} \bar{f_6} \bar{f_7} \bar{f_8} $ & $10101000010--00100000--------------------------$ \\ \hline
$\tau_{17}$ & $ a_1 \bar{a_2} {b_1} \bar{b_2} c_1 \bar{c_2} \bar{d_1} \bar{d_2} \bar{d_3} \bar{d_4} {d_5} \bar{f_1} \bar{f_2} {f_3} \bar{f_4} \bar{f_5} \bar{f_6} \bar{f_7} \bar{f_8} $ & $10101000001--00100000--------------------------$ \\ \hline
$\tau_{18}$ & $ a_1 \bar{a_2} {b_1} \bar{b_2} c_1 \bar{c_2} \bar{d_1} \bar{d_2} \bar{d_3} {d_4} \bar{d_5} \bar{f_1} \bar{f_2} \bar{f_3} {f_4} \bar{f_5} \bar{f_6} \bar{f_7} \bar{f_8} $ & $10101000010--00010000--------------------------$ \\ \hline
$\tau_{19}$ & $ a_1 \bar{a_2} {b_1} \bar{b_2} c_1 \bar{c_2} \bar{d_1} \bar{d_2} \bar{d_3} \bar{d_4} {d_5} \bar{f_1} \bar{f_2} \bar{f_3} {f_4} \bar{f_5} \bar{f_6} \bar{f_7} \bar{f_8} $ & $10101000001--00010000--------------------------$ \\ \hline
$\tau_{20}$ & $ a_1 \bar{a_2} {b_1} \bar{b_2} c_1 \bar{c_2} \bar{d_1} \bar{d_2} \bar{d_3} {d_4} \bar{d_5} \bar{f_1} \bar{f_2} \bar{f_3} \bar{f_4} {f_5} \bar{f_6} \bar{f_7} \bar{f_8} $ & $10101000010--00001000--------------------------$  \\ \hline
$\tau_{21}$ & $ a_1 \bar{a_2} {b_1} \bar{b_2} c_1 \bar{c_2} \bar{d_1} \bar{d_2} \bar{d_3} \bar{d_4} {d_5} \bar{f_1} \bar{f_2} \bar{f_3} \bar{f_4} {f_5} \bar{f_6} \bar{f_7} \bar{f_8} $ & $10101000001--00001000--------------------------$  \\ \hline

$\tau_{22}$ & $ a_1 \bar{a_2} {b_1} \bar{b_2} \bar{c_1} {c_2} {g_1} \bar{g_2} \bar{g_3} \bar{g_4} \bar{g_5} \bar{g_6} $ & $101001---------------100000--------------------$  \\ \hline
$\tau_{23}$ & $ a_1 \bar{a_2} {b_1} \bar{b_2} \bar{c_1} {c_2} \bar{g_1} {g_2} \bar{g_3} \bar{g_4} \bar{g_5} \bar{g_6} $ & $101001---------------010000--------------------$  \\ \hline
$\tau_{24}$ & $ a_1 \bar{a_2} {b_1} \bar{b_2} \bar{c_1} {c_2} \bar{g_1} \bar{g_2} {g_3} \bar{g_4} \bar{g_5} \bar{g_6} $ & $101001---------------001000--------------------$  \\ \hline
$\tau_{25}$ & $ a_1 \bar{a_2} {b_1} \bar{b_2} \bar{c_1} {c_2} \bar{g_1} \bar{g_2} \bar{g_3} {g_4} \bar{g_5} \bar{g_6} $ & $101001---------------000100--------------------$  \\ \hline
$\tau_{26}$ & $ a_1 \bar{a_2} {b_1} \bar{b_2} \bar{c_1} {c_2} \bar{g_1} \bar{g_2} \bar{g_3} \bar{g_4} {g_5} \bar{g_6} $ & $101001---------------000010--------------------$  \\ \hline

$\tau_{27}$ & $ a_1 \bar{a_2} \bar{g_1} \bar{g_2} {g_3} \bar{g_4} \bar{g_5} \bar{g_6} {i_1} \bar{i_2} \bar{i_3} {j_1} \bar{j_2} \bar{j_3} {k_1} \bar{k_2} $ & $1001-----------------001000---1001010----------$   \\ \hline

$\tau_{28}$ & $ a_1 \bar{a_2} \bar{b_1} {b_2} {g_1} \bar{g_2} \bar{g_3} \bar{g_4} \bar{g_5} \bar{g_6} {h_1} \bar{h_2} \bar{h_3} \bar{j_1} {j_2} $ & $1001-----------------10000010010001------------$  \\ \hline
$\tau_{22}$ & $ a_1 \bar{a_2} \bar{b_1} {b_2} \bar{g_1} {g_2} \bar{g_3} \bar{g_4} \bar{g_5} \bar{g_6} {h_1} \bar{h_2} \bar{h_3} \bar{j_1} {j_2} $ & $1001-----------------01000010010001------------$  \\ \hline
$\tau_{23}$ & $ a_1 \bar{a_2} \bar{b_1} {b_2} \bar{g_1} \bar{g_2} {g_3} \bar{g_4} \bar{g_5} \bar{g_6} {h_1} \bar{h_2} \bar{h_3} \bar{j_1} {j_2} $ & $1001-----------------00100010010001------------$  \\ \hline

$\tau_{24}$ & $ a_1 \bar{a_2} \bar{b_1} {b_2} {d_1} \bar{d_2} \bar{d_3} \bar{d_4} \bar{d_5}  {f_1} \bar{f_2} \bar{f_3} \bar{f_4} \bar{f_5} \bar{f_6} \bar{f_7} \bar{f_8} \bar{i_1} {i_2} \bar{i_3} \bar{m_1} {m_2} $ & $1001--10000--10000000---------010------01------$  \\ \hline
$\tau_{25}$ & $ a_1 \bar{a_2} \bar{b_1} {b_2} \bar{d_1} {d_2} \bar{d_3} \bar{d_4} \bar{d_5}  {f_1} \bar{f_2} \bar{f_3} \bar{f_4} \bar{f_5} \bar{f_6} \bar{f_7} \bar{f_8} \bar{i_1} {i_2} \bar{i_3} \bar{m_1} {m_2} $ & $1001--01000--10000000---------010------01------$  \\ \hline
$\tau_{26}$ & $ a_1 \bar{a_2} \bar{b_1} {b_2} \bar{d_1} \bar{d_2} {d_3} \bar{d_4} \bar{d_5}  {f_1} \bar{f_2} \bar{f_3} \bar{f_4} \bar{f_5} \bar{f_6} \bar{f_7} \bar{f_8} \bar{i_1} {i_2} \bar{i_3} \bar{m_1} {m_2} $ & $1001--00100--10000000---------010------01------$  \\ \hline
$\tau_{27}$ & $ a_1 \bar{a_2} \bar{b_1} {b_2} \bar{d_1} \bar{d_2} \bar{d_3} {d_4} \bar{d_5}  {f_1} \bar{f_2} \bar{f_3} \bar{f_4} \bar{f_5} \bar{f_6} \bar{f_7} \bar{f_8} \bar{i_1} {i_2} \bar{i_3} \bar{m_1} {m_2} $ & $1001--00010--10000000---------010------01------$  \\ \hline
$\tau_{28}$ & $ a_1 \bar{a_2} \bar{b_1} {b_2} {d_1} \bar{d_2} \bar{d_3} \bar{d_4} \bar{d_5}  \bar{f_1} {f_2} \bar{f_3} \bar{f_4} \bar{f_5} \bar{f_6} \bar{f_7} \bar{f_8} \bar{i_1} {i_2} \bar{i_3} \bar{m_1} {m_2} $ & $1001--10000--01000000---------010------01------$  \\ \hline
$\tau_{29}$ & $ a_1 \bar{a_2} \bar{b_1} {b_2} \bar{d_1} {d_2} \bar{d_3} \bar{d_4} \bar{d_5}  \bar{f_1} {f_2} \bar{f_3} \bar{f_4} \bar{f_5} \bar{f_6} \bar{f_7} \bar{f_8} \bar{i_1} {i_2} \bar{i_3} \bar{m_1} {m_2} $  & $1001--01000--01000000---------010------01------$  \\ \hline
$\tau_{30}$ & $ a_1 \bar{a_2} \bar{b_1} {b_2} \bar{d_1} \bar{d_2} {d_3} \bar{d_4} \bar{d_5}  \bar{f_1} {f_2} \bar{f_3} \bar{f_4} \bar{f_5} \bar{f_6} \bar{f_7} \bar{f_8} \bar{i_1} {i_2} \bar{i_3} \bar{m_1} {m_2} $  & $1001--00100--01000000---------010------01------$  \\ \hline
$\tau_{31}$ & $ a_1 \bar{a_2} \bar{b_1} {b_2} \bar{d_1} \bar{d_2} \bar{d_3} {d_4} \bar{d_5}  \bar{f_1} {f_2} \bar{f_3} \bar{f_4} \bar{f_5} \bar{f_6} \bar{f_7} \bar{f_8} \bar{i_1} {i_2} \bar{i_3} \bar{m_1} {m_2} $  & $1001--00010--01000000---------010------01------$ \\ \hline
$\tau_{32}$ & $ a_1 \bar{a_2} \bar{b_1} {b_2} {d_1} \bar{d_2} \bar{d_3} \bar{d_4} \bar{d_5}  \bar{f_1} \bar{f_2} {f_3} \bar{f_4} \bar{f_5} \bar{f_6} \bar{f_7} \bar{f_8} \bar{i_1} {i_2} \bar{i_3} \bar{m_1} {m_2} $  & $1001--10000--00100000---------010------01------$  \\ \hline
$\tau_{33}$ & $ a_1 \bar{a_2} \bar{b_1} {b_2} \bar{d_1} {d_2} \bar{d_3} \bar{d_4} \bar{d_5}  \bar{f_1} \bar{f_2} {f_3} \bar{f_4} \bar{f_5} \bar{f_6} \bar{f_7} \bar{f_8} \bar{i_1} {i_2} \bar{i_3} \bar{m_1} {m_2} $ & $1001--01000--00100000---------010------01------$ \\ \hline
$\tau_{34}$ & $ a_1 \bar{a_2} \bar{b_1} {b_2} \bar{d_1} \bar{d_2} {d_3} \bar{d_4} \bar{d_5}  \bar{f_1} \bar{f_2} {f_3} \bar{f_4} \bar{f_5} \bar{f_6} \bar{f_7} \bar{f_8} \bar{i_1} {i_2} \bar{i_3} \bar{m_1} {m_2} $ & $1001--00100--00100000---------010------01------$ \\ \hline
$\tau_{35}$ & $ a_1 \bar{a_2} \bar{b_1} {b_2} \bar{d_1} \bar{d_2} \bar{d_3} {d_4} \bar{d_5}  \bar{f_1} \bar{f_2} {f_3} \bar{f_4} \bar{f_5} \bar{f_6} \bar{f_7} \bar{f_8} \bar{i_1} {i_2} \bar{i_3} \bar{m_1} {m_2} $ & $1001--00010--00100000---------010------01------$ \\ \hline

$\tau_{36}$ & $ a_1 \bar{a_2} \bar{b_1} {b_2} {d_1} \bar{d_2} \bar{d_3} \bar{d_4} \bar{d_5} \bar{f_1} \bar{f_2} \bar{f_3} {f_4} \bar{f_5} \bar{f_6} \bar{f_7} \bar{f_8} \bar{i_1} {i_2} \bar{i_3} $ & $1001--10000--00010000---------010--------------$   \\ \hline
$\tau_{37}$ & $ a_1 \bar{a_2} \bar{b_1} {b_2} \bar{d_1} {d_2} \bar{d_3} \bar{d_4} \bar{d_5} \bar{f_1} \bar{f_2} \bar{f_3} {f_4} \bar{f_5} \bar{f_6} \bar{f_7} \bar{f_8} \bar{i_1} {i_2} \bar{i_3} $ & $1001--01000--00010000---------010--------------$ \\ \hline
$\tau_{38}$ & $ a_1 \bar{a_2} \bar{b_1} {b_2} \bar{d_1} \bar{d_2} {d_3} \bar{d_4} \bar{d_5} \bar{f_1} \bar{f_2} \bar{f_3} {f_4} \bar{f_5} \bar{f_6} \bar{f_7} \bar{f_8} \bar{i_1} {i_2} \bar{i_3} $ & $1001--00100--00010000---------010--------------$ \\ \hline
$\tau_{39}$ & $ a_1 \bar{a_2} \bar{b_1} {b_2} \bar{d_1} \bar{d_2} \bar{d_3} {d_4} \bar{d_5} \bar{f_1} \bar{f_2} \bar{f_3} {f_4} \bar{f_5} \bar{f_6} \bar{f_7} \bar{f_8} \bar{i_1} {i_2} \bar{i_3} $ & $1001--00010--00010000---------010--------------$  \\ \hline

\end{tabular}
\egroup
}
\end{table*}

\begin{table*}[!htbp]
\scriptsize
\centering
\caption{Prime Implicants explanations for DDoS LOIC HTTP attack (2/2).}
\label{table_loichttp2}
{
\bgroup
\def\arraystretch{1.5}
\begin{tabular}{|c|c|c|}
\hline
\# &Minterm & Boolean Expression \\ \hline

$\tau_{40}$ & $ a_1 \bar{a_2} \bar{b_1} {b_2} {d_1} \bar{d_2} \bar{d_3} \bar{d_4} \bar{d_5} \bar{f_1} \bar{f_2} \bar{f_3} {f_4} \bar{f_5} \bar{f_6} \bar{f_7} \bar{f_8} \bar{i_1} \bar{i_2} {i_3} $ & $1001--10000--00010000---------001--------------$ \\ \hline
$\tau_{41}$ & $ a_1 \bar{a_2} \bar{b_1} {b_2} \bar{d_1} {d_2} \bar{d_3} \bar{d_4} \bar{d_5} \bar{f_1} \bar{f_2} \bar{f_3} {f_4} \bar{f_5} \bar{f_6} \bar{f_7} \bar{f_8} \bar{i_1} \bar{i_2} {i_3} $ & $1001--01000--00010000---------001--------------$ \\ \hline
$\tau_{42}$ & $ a_1 \bar{a_2} \bar{b_1} {b_2} \bar{d_1} \bar{d_2} {d_3} \bar{d_4} \bar{d_5} \bar{f_1} \bar{f_2} \bar{f_3} {f_4} \bar{f_5} \bar{f_6} \bar{f_7} \bar{f_8} \bar{i_1} \bar{i_2} {i_3} $ & $1001--00100--00010000---------001--------------$ \\ \hline
$\tau_{43}$ & $ a_1 \bar{a_2} \bar{b_1} {b_2} \bar{d_1} \bar{d_2} \bar{d_3} {d_4} \bar{d_5} \bar{f_1} \bar{f_2} \bar{f_3} {f_4} \bar{f_5} \bar{f_6} \bar{f_7} \bar{f_8} \bar{i_1} \bar{i_2} {i_3} $ & $1001--00010--00010000---------001--------------$  \\ \hline

$\tau_{44}$ & $ a_1 \bar{a_2} \bar{b_1} {b_2} \bar{d_1} \bar{d_2} \bar{d_3} \bar{d_4} {d_5}  {g_1} \bar{g_2} \bar{g_3} \bar{g_4} \bar{g_5} \bar{g_6} \bar{i_1} {i_2} \bar{i_3} \bar{n_1} {n_2} \bar{o_1} {o_2} $ & $1001--00001----------100000---010--------0101--$ \\ \hline
$\tau_{45}$ & $ a_1 \bar{a_2} \bar{b_1} {b_2} \bar{d_1} \bar{d_2} \bar{d_3} \bar{d_4} {d_5}  \bar{g_1} {g_2} \bar{g_3} \bar{g_4} \bar{g_5} \bar{g_6} \bar{i_1} {i_2} \bar{i_3} \bar{n_1} {n_2} \bar{o_1} {o_2} $ & $1001--00001----------010000---010--------0101--$ \\ \hline
$\tau_{46}$ & $ a_1 \bar{a_2} \bar{b_1} {b_2} \bar{d_1} \bar{d_2} \bar{d_3} \bar{d_4} {d_5}  \bar{g_1} \bar{g_2} {g_3} \bar{g_4} \bar{g_5} \bar{g_6} \bar{i_1} {i_2} \bar{i_3} \bar{n_1} {n_2} \bar{o_1} {o_2} $ & $1001--00001----------001000---010--------0101--$ \\ \hline
$\tau_{47}$ & $ a_1 \bar{a_2} \bar{b_1} {b_2} \bar{d_1} \bar{d_2} \bar{d_3} \bar{d_4} {d_5}  \bar{g_1} \bar{g_2} \bar{g_3} {g_4} \bar{g_5} \bar{g_6} \bar{i_1} {i_2} \bar{i_3} \bar{n_1} {n_2} \bar{o_1} {o_2} $ & $1001--00001----------000100---010--------0101--$ \\ \hline
$\tau_{48}$ & $ a_1 \bar{a_2} \bar{b_1} {b_2} \bar{d_1} \bar{d_2} \bar{d_3} \bar{d_4} {d_5}  {g_1} \bar{g_2} \bar{g_3} \bar{g_4} \bar{g_5} \bar{g_6} \bar{i_1} \bar{i_2} {i_3} \bar{n_1} {n_2} \bar{o_1} {o_2} $ & $1001--00001----------100000---001--------0101--$ \\ \hline
$\tau_{49}$ & $ a_1 \bar{a_2} \bar{b_1} {b_2} \bar{d_1} \bar{d_2} \bar{d_3} \bar{d_4} {d_5}  \bar{g_1} {g_2} \bar{g_3} \bar{g_4} \bar{g_5} \bar{g_6} \bar{i_1} \bar{i_2} {i_3} \bar{n_1} {n_2} \bar{o_1} {o_2} $ & $1001--00001----------010000---001--------0101--$  \\ \hline
$\tau_{50}$ & $ a_1 \bar{a_2} \bar{b_1} {b_2} \bar{d_1} \bar{d_2} \bar{d_3} \bar{d_4} {d_5}  \bar{g_1} \bar{g_2} {g_3} \bar{g_4} \bar{g_5} \bar{g_6} \bar{i_1} \bar{i_2} {i_3} \bar{n_1} {n_2} \bar{o_1} {o_2} $ & $1001--00001----------001000---001--------0101--$ \\ \hline
$\tau_{51}$ & $ a_1 \bar{a_2} \bar{b_1} {b_2} \bar{d_1} \bar{d_2} \bar{d_3} \bar{d_4} {d_5}  \bar{g_1} \bar{g_2} \bar{g_3} {g_4} \bar{g_5} \bar{g_6} \bar{i_1} \bar{i_2} {i_3} \bar{n_1} {n_2} \bar{o_1} {o_2} $ & $1001--00001----------000100---001--------0101--$ \\ \hline

$\tau_{52}$ & $ \bar{a_1} {a_2} \bar{p_1} {p_2}$ & $01-------------------------------------------01$ \\ \hline

\end{tabular}
\egroup
}
\end{table*}

\begin{table*}[!htbp]
\scriptsize
\centering
\caption{Prime Implicants explanations for DDoS LOIC UDP attack.}
\label{table_loicudp}
{
\bgroup
\def\arraystretch{1.5}
\begin{tabular}{|c|c|c|}
\hline
\# &Minterm & Boolean Expression \\ \hline

$\tau_{53}$ & $ {d_1} \bar{d_2} \bar{d_3} \bar{d_4} \bar{d_5} {f_1} \bar{f_2} \bar{f_3} \bar{f_4} \bar{f_5} \bar{f_6} \bar{f_7} \bar{f_8} {g_1} \bar{g_2} \bar{g_3} $ & $------10000--10000000100000--------------------$ \\ \hline

$\tau_{54}$ & $ {d_1} \bar{d_2} \bar{d_3} \bar{d_4} \bar{d_5} \bar{f_1} {f_2} \bar{f_3} \bar{f_4} \bar{f_5} \bar{f_6} \bar{f_7} \bar{f_8} {g_1} \bar{g_2} \bar{g_3} $ & $------10000--01000000100000--------------------$ \\ \hline

$\tau_{55}$ & $ {d_1} \bar{d_2} \bar{d_3} \bar{d_4} \bar{d_5} \bar{f_1} \bar{f_2} {f_3} \bar{f_4} \bar{f_5} \bar{f_6} \bar{f_7} \bar{f_8} {g_1} \bar{g_2} \bar{g_3} $ & $------10000--00100000100000--------------------$ \\ \hline

$\tau_{56}$ & $ {d_1} \bar{d_2} \bar{d_3} \bar{d_4} \bar{d_5} \bar{f_1} \bar{f_2} \bar{f_3} {f_4} \bar{f_5} \bar{f_6} \bar{f_7} \bar{f_8} {g_1} \bar{g_2} \bar{g_3} $ & $------10000--00010000100000--------------------$ \\ \hline

$\tau_{57}$ & $ {d_1} \bar{d_2} \bar{d_3} \bar{d_4} \bar{d_5} \bar{f_1} \bar{f_2} \bar{f_3} \bar{f_4} {f_5} \bar{f_6} \bar{f_7} \bar{f_8} {g_1} \bar{g_2} \bar{g_3} $ & $------10000--00001000100000--------------------$ \\ \hline

$\tau_{58}$ & $ {d_1} \bar{d_2} \bar{d_3} \bar{d_4} \bar{d_5} \bar{f_1} \bar{f_2} \bar{f_3} \bar{f_4} \bar{f_5} {f_6} \bar{f_7} \bar{f_8} {g_1} \bar{g_2} \bar{g_3} $ & $------10000--00000100100000--------------------$ \\ \hline

$\tau_{59}$ & $ {d_1} \bar{d_2} \bar{d_3} \bar{d_4} \bar{d_5} \bar{f_1} \bar{f_2} \bar{f_3} \bar{f_4} \bar{f_5} \bar{f_6} {f_7} \bar{f_8} {g_1} \bar{g_2} \bar{g_3} $ & $------10000--00000010100000--------------------$ \\ \hline

$\tau_{60}$ & $ {d_1} \bar{d_2} \bar{d_3} \bar{d_4} \bar{d_5} {f_1} \bar{f_2} \bar{f_3} \bar{f_4} \bar{f_5} \bar{f_6} \bar{f_7} \bar{f_8} \bar{g_1} {g_2} \bar{g_3} $ & $------10000--10000000010000--------------------$ \\ \hline

$\tau_{61}$ & $ {d_1} \bar{d_2} \bar{d_3} \bar{d_4} \bar{d_5} \bar{f_1} {f_2} \bar{f_3} \bar{f_4} \bar{f_5} \bar{f_6} \bar{f_7} \bar{f_8} \bar{g_1} {g_2} \bar{g_3} $ & $------10000--01000000010000--------------------$ \\ \hline

$\tau_{62}$ & $ {d_1} \bar{d_2} \bar{d_3} \bar{d_4} \bar{d_5} \bar{f_1} \bar{f_2} {f_3} \bar{f_4} \bar{f_5} \bar{f_6} \bar{f_7} \bar{f_8} \bar{g_1} {g_2} \bar{g_3} $ & $------10000--00100000010000--------------------$ \\ \hline

$\tau_{63}$ & $ {d_1} \bar{d_2} \bar{d_3} \bar{d_4} \bar{d_5} \bar{f_1} \bar{f_2} \bar{f_3} {f_4} \bar{f_5} \bar{f_6} \bar{f_7} \bar{f_8} \bar{g_1} {g_2} \bar{g_3} $ & $------10000--00010000010000--------------------$ \\ \hline

$\tau_{64}$ & $ {d_1} \bar{d_2} \bar{d_3} \bar{d_4} \bar{d_5} \bar{f_1} \bar{f_2} \bar{f_3} \bar{f_4} {f_5} \bar{f_6} \bar{f_7} \bar{f_8} \bar{g_1} {g_2} \bar{g_3} $ & $------10000--00001000010000--------------------$ \\ \hline

$\tau_{65}$ & $ {d_1} \bar{d_2} \bar{d_3} \bar{d_4} \bar{d_5} \bar{f_1} \bar{f_2} \bar{f_3} \bar{f_4} \bar{f_5} {f_6} \bar{f_7} \bar{f_8} \bar{g_1} {g_2} \bar{g_3} $ & $------10000--00000100010000--------------------$ \\ \hline

$\tau_{66}$ & $ {d_1} \bar{d_2} \bar{d_3} \bar{d_4} \bar{d_5} \bar{f_1} \bar{f_2} \bar{f_3} \bar{f_4} \bar{f_5} \bar{f_6} {f_7} \bar{f_8} \bar{g_1} {g_2} \bar{g_3} $ & $------10000--00000010010000--------------------$ \\ \hline

$\tau_{60}$ & $ {d_1} \bar{d_2} \bar{d_3} \bar{d_4} \bar{d_5} {f_1} \bar{f_2} \bar{f_3} \bar{f_4} \bar{f_5} \bar{f_6} \bar{f_7} \bar{f_8} \bar{g_1} \bar{g_2} {g_3} $ & $------10000--10000000001000--------------------$ \\ \hline

$\tau_{61}$ & $ {d_1} \bar{d_2} \bar{d_3} \bar{d_4} \bar{d_5} \bar{f_1} {f_2} \bar{f_3} \bar{f_4} \bar{f_5} \bar{f_6} \bar{f_7} \bar{f_8} \bar{g_1} \bar{g_2} {g_3} $ & $------10000--01000000001000--------------------$ \\ \hline

$\tau_{62}$ & $ {d_1} \bar{d_2} \bar{d_3} \bar{d_4} \bar{d_5} \bar{f_1} \bar{f_2} {f_3} \bar{f_4} \bar{f_5} \bar{f_6} \bar{f_7} \bar{f_8} \bar{g_1} \bar{g_2} {g_3} $ & $------10000--00100000001000--------------------$ \\ \hline

$\tau_{63}$ & $ {d_1} \bar{d_2} \bar{d_3} \bar{d_4} \bar{d_5} \bar{f_1} \bar{f_2} \bar{f_3} {f_4} \bar{f_5} \bar{f_6} \bar{f_7} \bar{f_8} \bar{g_1} \bar{g_2} {g_3} $ & $------10000--00010000001000--------------------$ \\ \hline

$\tau_{64}$ & $ {d_1} \bar{d_2} \bar{d_3} \bar{d_4} \bar{d_5} \bar{f_1} \bar{f_2} \bar{f_3} \bar{f_4} {f_5} \bar{f_6} \bar{f_7} \bar{f_8} \bar{g_1} \bar{g_2} {g_3} $ & $------10000--00001000001000--------------------$ \\ \hline

$\tau_{65}$ & $ {d_1} \bar{d_2} \bar{d_3} \bar{d_4} \bar{d_5} \bar{f_1} \bar{f_2} \bar{f_3} \bar{f_4} \bar{f_5} {f_6} \bar{f_7} \bar{f_8} \bar{g_1} \bar{g_2} {g_3} $ & $------10000--00000100001000--------------------$ \\ \hline

$\tau_{66}$ & $ {d_1} \bar{d_2} \bar{d_3} \bar{d_4} \bar{d_5} \bar{f_1} \bar{f_2} \bar{f_3} \bar{f_4} \bar{f_5} \bar{f_6} {f_7} \bar{f_8} \bar{g_1} \bar{g_2} {g_3} $ & $------10000--00000010001000--------------------$ \\ \hline
\end{tabular}
\egroup
}
\end{table*}

\begin{table*}[!htbp]
\scriptsize
\centering
\caption{Prime Implicants explanations for DDoS HOIC attack.}
\label{table_hoic}
{
\bgroup
\def\arraystretch{1.5}
\begin{tabular}{|c|c|c|}
\hline
\# &Minterm & Boolean Expression \\ \hline

$\tau_{67}$ & $ \bar{f_1} \bar{f_2} \bar{f_3} \bar{f_4} \bar{f_5} \bar{f_6} \bar{f_7} {f_8} \bar{g_1} {g_2} \bar{g_3} $ & $-------------00000001010000--------------------$ \\ \hline

$\tau_{68}$ & $ \bar{f_1} \bar{f_2} \bar{f_3} \bar{f_4} \bar{f_5} \bar{f_6} \bar{f_7} {f_8} \bar{g_1} \bar{g_2} {g_3} $ & $-------------00000001001000--------------------$ \\ \hline
\end{tabular}
\egroup
}
\end{table*}

For example, a flow instance with the features shown in Algorithm \ref{alg1}, will be mapped into 
$$a_1\bar{a_2}{b_1}\bar{b_2}\bar{c_1}{c_2}{d_1}\bar{d_2}\bar{d_3}\bar{d_4}\bar{d_5}{e_1}\bar{e_2}\bar{f_1}\bar{f_2}\bar{f_3}\bar{f_4}\bar{f_5}\bar{f_6}\bar{f_7}{f_8}\bar{g_1}\bar{g_2}\bar{g_3}$$$$\bar{g_4}{g_5}\bar{g_6}{h_1}\bar{h_2}\bar{h_3}{i_1}\bar{i_2}\bar{i_3}\bar{j_1}{j_2}{k_1}\bar{k_2}{l_1}\bar{l_2}{m_1}\bar{m_2}{n_1}\bar{n_2}{o_1}\bar{o_2}{p_1}\bar{p_2}$$
which can be represented in numeric:

$$10100110000100000000100001010010001101010101010$$

\subsection{Prime Implicant Explanations}

After the M\&M feature discretization, the model can be represented into boolean expressions, and prime Implicants are calculated with Quine-McCluskey algorithm. The prime implicants explanations for \emph{DDoS LOIC HTTP}, \emph{DDoS LOIC UDP}, and \emph{DDoS HOIC} in minterms are shown in Table \ref{table_loichttp1}, \ref{table_loichttp2}, \ref{table_loicudp}, and \ref{table_hoic}, in which a ``$-$'' means a ``don't care''. The behavior of the DDoS traffic detection of the decision tree model can be rigorously interpreted as $$\Delta = {\tau_1}\vee{\tau_2}\vee \cdots \vee {\tau_{68}}$$

\emph{Prime implicants explanation} examples for five real-life benign traffic flow instances explanation is presented below. The original features, discretized features, and boolean expression of each instance are presented in detail. The reason for why it is classified as benign is provided and marked in red in its boolean expression, by the prime implicant that is used to make the decision, which is sufficient and rigorous to explain the decision. 

\textbf{\emph{Use Case 1:}} In flow instance 1, its boolean expression is 
$$0011000010010100100100101000101100110$$ 
after feature discretization and mapping, and it matches with \emph{prime implicant $\tau_{41}$}: 

\begin{small}
$$001--------------------------------10$$
\end{small}

which means \emph{``if a flow's boolean expression start with $001$, and end with $10$, no matter what values is for the other bits, it is a benign flow.''}. With the feature discretization mapping method, \emph{prime implicant $\tau_{41}$} is originally ``$\bar{a_1} \bar{a_2} {a_3} {n_1} \bar{n_2}$'' which can also be interpreted as \emph{``if a flow's Fwd\_Pkt\_Len\_Min is larger than $110.5$, and Fwd\_Seg\_Size\_Avg is smaller than $486$, then it is benign.''} Thus, the model decides this flow is benign. The whole decision process is formal and rigorous, and the reasons is sufficient, as proved in Section \ref{sec_xai}. 

\textbf{\emph{Use Case 2:}} In flow instance 5, whose feature is mapped into boolean expression $$0100100100010100100100101010001100110$$ is classified into ``Benign'' because it matches with \emph{prime implicant $\tau_{39}$: } 

\begin{small}
$$0100100--------------------------01--$$
\end{small}
which is $$\bar{a_1} {a_2} \bar{a_3} \bar{b_1} {b_2} \bar{b_3} \bar{b_4} \bar{m_1} {m_2}$$ says \emph{``if a flow has Fwd\_Pkt\_Len\_Min between $49.5$ and $110.5$, Fwd\_Pkt\_s between $0.01$ and $3793$, and Fwd\_IAT\_Tot larger than $486$, then it is benign.''} It is worth noting that one instance can match more than one (sometimes even many) \emph{prime implicants} (although we did not have such experience in this experiment), which means that there are multiple explanations for the decision made on that instance, and each one of these reasons is sufficient and rigorous. As the explanation method is based on formal logic, the rules explained is guaranteed to have the same behavior with the model, thus the accuracy of \sys~is the same as the decision tree model, which is compared with other models in Tables. \ref{table_hoic} and \ref{table_loic_udp}. As shown in Tables \ref{table_hoic} and \ref{table_loic_udp}, the true-positive rate for all DDoS intrusions discussed in this paper are $100\%$, and the false-positive rate are all $0\%$.

\begin{table*}[ht]
\tiny
\centering
\caption{Results of DDoS-HOIC Attack on \sys~and other Models. }
\label{table_hoic}
\resizebox{\textwidth}{11mm}{
\begin{tabular}{|c|c|c|c|c|c|c|c|}
\hline
\multicolumn{2}{|c|}{} &\textbf{Random Forest} & \textbf{Naive Bayes} & {\color{red}{ \textbf{\sys} }}& \textbf{Neural Network (MLP)} & \textbf{Quadratic Discriminante} & \textbf{KNeighbors} \\ \hline
\multicolumn{2}{|c|}{\textbf{True-Positive}} & 100\% & 100\% & {\color{red}{100\%}} & 100\% & 100\% & 100\% \\ \hline
\multicolumn{2}{|c|}{\textbf{False-Positive}} & 0\% & 0\% & {\color{red}{0\%}} & 0\% & 0\% & 0\% \\ \hline
\multirow{2}*{\textbf{recall}} & \textbf{Benigh} & 1.0 & 1.0 & {\color{red}{1.0}} & 1.0 & 1.0 & 1.0 \\ \cline{2-8}
& \textbf{DDoS-HOIC} & 1.0 & 1.0 & {\color{red}{1.0}} & 1.0 & 1.0 & 1.0 \\  \hline
\multirow{2}*{\textbf{f1-score}} & \textbf{Benigh} & 1.0 & 1.0 & {\color{red}{1.0}} & 1.0 & 1.0 & 1.0 \\ \cline{2-8}
& \textbf{DDoS-HOIC} & 1.0 & 1.0 & {\color{red}{1.0}} & 1.0 & 1.0 & 1.0 \\  \hline
\end{tabular}
}
\end{table*}

\begin{table*}[ht]
\tiny
\centering
\caption{Results of DDOS-LOIC-UDP Attack on \sys~and other Models.}
\label{table_loic_udp}
\resizebox{\textwidth}{11mm}{
\begin{tabular}{|c|c|c|c|c|c|c|c|}
\hline
\multicolumn{2}{|c|}{} &\textbf{Random Forest} & \textbf{Naive Bayes} & {\color{red}{\textbf{\sys}}} & \textbf{Neural Network (MLP)} & \textbf{Quadratic Discriminante} & \textbf{KNeighbors} \\ \hline
\multicolumn{2}{|c|}{\textbf{True-Positive}} & 100\% & 100\% & {\color{red}{100\%}} & 100\% & 100\% & 100\% \\ \hline
\multicolumn{2}{|c|}{\textbf{False-Positive}} & 0\% & 0\% & {\color{red}{0\%}} & 0\% & 0\% & 0\% \\ \hline
\multirow{2}*{\textbf{recall}} & \textbf{Benigh} & 1.0 & 1.0 & {\color{red}{1.0}} & 1.0 & 1.0 & 1.0 \\ \cline{2-8}
& \textbf{DDOS-LOIC-UDP} & 1.0 & 0.66 & {\color{red}{1.0}} & 0.93 & 0.66 & 0.69 \\  \hline
\multirow{2}*{\textbf{f1-score}} & \textbf{Benigh} & 1.0 & 1.0 & {\color{red}{1.0}} & 1.0 & 1.0 & 1.0 \\ \cline{2-8}
& \textbf{DDOS-LOIC-UDP} & 1.0 & 0.80 & {\color{red}{1.0}} & 0.97 & 0.80 & 0.81 \\  \hline
\end{tabular}
}
\end{table*}

\section{Conclusion and future research challenges}\label{sec_sum}

The implementation of an machine learning driven intrusion detection system depends entirely on the ability to explicitly and sufficiently interpret the machine learning models. However, current machine learning interpretation methods in intrusion detection are heuristic, which could not garrantee the accracy or sufficiency of the rules explained. In this paper, we have proposed a rigorous rules extraction method for identifying DDoS traffic flow from a decision tree model with 100\% accuracy. By discretizing the continuous model into boolean expression and calculating the prime implicants out of it, the proposed map, combine, and merge method is able to provide sufficient and rigorous explanations for DDoS detection. As the \sys~method is based on formal logic calculation, the rules extracted have exactly the same behavior with the model.

Although a rigorous XAI driven AIS have been proposed for the first time in our humble knowledge, several limitations remain and are challenging. 

-- First, the rule extraction process is built based on the hypothesis that features used are independent with each other. However, it may not hold in real life. How to revise the features to make sure they are independent require expert knowledge and experience and therefore more work. 

-- Second, the rigorous XAI technology depend on prime implicants, and the prime implicant calculation methodology, although being under heated discussion and more tools keep emerging, is not guaranteed to finish within polynomial time and space complexity. The time and space expense highly depend on the boolean expression to solve.  

-- Third, the extraction of rules from a well-performed machine learning model is not the end of story. For the model keeps changing each time it is fitted with new data, and the rules extracted from it are not the same every time. It is a controversial to claim which one is the right one, for the machine learning algorithms can only recognize patterns from data, but cannot tell which one is the correct one that will hold in the future (due to Hume's law). Human experts should be involved in to use their intelligent and experience to work out the final rules for benign traffic detection.

-- Finally, although in this paper, I explained decision tree model for it performs better than the other models in detecting known intrusions according to previous work, it is possible that rigorous explanation for other models are required in other scenarios. As the features in intrusion detection domain are usually continuous, the discretization, boolean expression generation, and prime implicants calculation for other machine learning models (such as deep learning, SVM, etc.) are much more challenging than decision tree, and much work still remain to be done.

To summarize, although a promising step towards rigorous XAI driven DDoS intrusion detection system has been made, still much is to be done to make it practical and scalable, especially human experts' effort are required and progress on SAT solver will also means significantly towards this goal.

\begin{table*}[!htbp]
\scriptsize
\centering
\caption{Examples of Prime Implicant Explanations for Benign Flow Instance.}
\label{table_eval}
\resizebox{\textwidth}{5cm}
{
\bgroup
\def\arraystretch{1.5}
\begin{tabular}{|c|c|c|c|c|c|c|c|c|c|c|c|c|c|c|}
\hline
&A: & B: & C: & D: & E: & F: & G: & H: & I: & J: & K: & L: & M: & N: \\ \hline 

\multirow{3}*{1} & 146& 0.0 & 30043443.7 & 90130331& 9627.8 & 4& 32& 30051116& 146& 17500& 30032640& 30032640& 90130331& 146 \\ \cline{2-15}

& \multicolumn{14}{c|}{$\bar{a_1} \bar{a_2} {a_3} {b_1} \bar{b_2} \bar{b_3} \bar{b_4} \bar{c_1} {c_2} \bar{c_3} \bar{d_1} {d_2} \bar{d_3} {e_1} \bar{e_2} \bar{e_3} {f_1} \bar{f_2} \bar{f_3} {g_1} \bar{g_2} \bar{h_1} {h_2} \bar{h_3} {i_1} \bar{i_2} \bar{j_1} \bar{j_2} {j_3} \bar{k_1} {k_2} {l_1} \bar{l_2} \bar{m_1} {m_2} {n_1} \bar{n_2}$} \\ \cline{2-15}

& \multicolumn{14}{c|}{{\color{red}{$001$}}$10000100101001001001010001011001${\color{red}{$10$}}} \\ \cline{1-15}

\multirow{2}*{PI $\tau_{41}:$} &\multicolumn{14}{c|}{$ 001--------------------------------10$} \\ \cline{2-15}

&\multicolumn{14}{c|}{$ \bar{a_1} \bar{a_2} {a_3} {n_1} \bar{n_2} $} \\ \cline{1-15} \cline{1-15} \cline{1-15} \hline

\multirow{3}*{2} & 0.0 & 1& 0.0 & 119999476& 130.9	& 121& 0& 1000244& 0.0 & 0.0 & 999705& 0.0 & 119999476& 0.0 \\ \cline{2-15}

& \multicolumn{14}{c|}{${a_1} \bar{a_2} \bar{a_3} \bar{b_1} {b_2} \bar{b_3} \bar{b_4} {c_1} \bar{c_2} \bar{c_3} \bar{d_1} \bar{d_2} {d_3} {e_1} \bar{e_2} \bar{e_3} \bar{f_1} {f_2} \bar{f_3} {g_1} \bar{g_2} \bar{h_1} {h_2} \bar{h_3} {i_1} \bar{i_2} {j_1} \bar{j_2} \bar{j_3} \bar{k_1} {k_2} {l_1} \bar{l_2} \bar{m_1} {m_2} {n_1} \bar{n_2}$} \\ \cline{2-15}

& \multicolumn{14}{c|}{{\color{red}{$1000100$}}$100${\color{red}{$001$}}$100${\color{red}{$010$}}$100101010001100110$} \\ \cline{1-15}

\multirow{2}*{PI $\tau_{36}:$} &\multicolumn{14}{c|}{$ 1000100---001---010------------------$} \\ \cline{2-15}
&\multicolumn{14}{c|}{$ a_1 \bar{a_2} \bar{a_3} \bar{b_1} {b_2} \bar{b_3} \bar{b_4} \bar{d_1} \bar{d_2} {d_3} \bar{f_1} {f_2} \bar{f_3} $} \\ \cline{1-15} \cline{1-15} \hline

\multirow{3}*{3} & 0& 0.2 & 45007702.5 & 90808764& 15355813.3	 & 17& 352& 45015090& 517& 443& 0& 45000315& 90803655& 67.7 \\ \cline{2-15}
& \multicolumn{14}{c|}{$\bar{a_1} \bar{a_2} {a_3} \bar{b_1} {b_2} \bar{b_3} \bar{b_4} \bar{c_1} {c_2} \bar{c_3} \bar{d_1} {d_2} \bar{d_3} \bar{e_1} {e_2} \bar{e_3} {f_1} \bar{f_2} \bar{f_3} {g_1} \bar{g_2} \bar{h_1} \bar{h_2} {h_3} {i_1} \bar{i_2} \bar{j_1} {j_2} \bar{j_3} {k_1} \bar{k_2} {l_1} \bar{l_2} \bar{m_1} {m_2} {n_1} \bar{n_2}$} \\ \cline{2-15}

& \multicolumn{14}{c|}{{\color{red}{$001$}}$0100100100101001000110010101001${\color{red}{$10$}}} \\ \cline{1-15}

\multirow{2}*{PI $\tau_{41}:$} &\multicolumn{14}{c|}{$ 001--------------------------------10$} \\ \cline{2-15}

&\multicolumn{14}{c|}{$ \bar{a_1} \bar{a_2} {a_3} {n_1} \bar{n_2} $} \\ \cline{1-15} \cline{1-15} \hline

\multirow{3}*{4} & 40& 0.0 & 38448540.3 & 115345621.0 & 358398.1 & 4& 32& 38687191& 40& 1947& 38036412 & 38036412& 115345621& 40 \\ \cline{2-15}

& \multicolumn{14}{c|}{${a_1} \bar{a_2} \bar{a_3} {b_1} \bar{b_2} \bar{b_3} \bar{b_4} \bar{c_1} {c_2} \bar{c_3} \bar{d_1} \bar{d_2} {d_3} {e_1} \bar{e_2} \bar{e_3} {f_1} \bar{f_2} \bar{f_3} {g_1} \bar{g_2} \bar{h_1} {h_2} \bar{h_3} {i_1} \bar{i_2} \bar{j_1} \bar{j_2} {j_3} \bar{k_1} \bar{k_2} {l_1} \bar{l_2} \bar{m_1} {m_2} {n_1} \bar{n_2}$} \\ \cline{2-15}

& \multicolumn{14}{c|}{{\color{red}{$1001000010$}}$010100100100101000110100110$} \\ \cline{1-15}

\multirow{2}*{PI $\tau_{1}:$} &\multicolumn{14}{c|}{$1001000010---------------------------$} \\ \cline{2-15}

&\multicolumn{14}{c|}{$a_1 \bar{a_2} \bar{a_3} b_1 \bar{b_2} \bar{b_3} \bar{b_4}\bar{c_1} c_2 \bar{c_3} $} \\ \cline{1-15} \cline{1-15} \hline

\multirow{3}*{5} & 50 & 2 & 0.0 & 1514340 & 7697.6 & 3.0 & 24 & 762613 & 50 & 137 & 751727 & 0.0 & 1514340 & 50 \\ \cline{2-15}

& \multicolumn{14}{c|}{$\bar{a_1} {a_2} \bar{a_3} \bar{b_1} {b_2} \bar{b_3} \bar{b_4} {c_1} \bar{c_2} \bar{c_3} \bar{d_1} {d_2} \bar{d_3} {e_1} \bar{e_2} \bar{e_3} {f_1} \bar{f_2} \bar{f_3} {g_1} \bar{g_2} \bar{h_1} {h_2} \bar{h_3} {i_1} \bar{i_2} {j_1} \bar{j_2} \bar{j_3} \bar{k_1} {k_2} {l_1} \bar{l_2} \bar{m_1} {m_2} {n_1} \bar{n_2}$} \\ \cline{2-15}

& \multicolumn{14}{c|}{{\color{red}{$0100100$}}$10001010010010010101000110${\color{red}{$01$}}$10$} \\ \cline{1-15}

\multirow{2}*{PI $\tau_{39}:$} &\multicolumn{14}{c|}{$0100100--------------------------01--$} \\ \cline{2-15}

&\multicolumn{14}{c|}{$\bar{a_1} {a_2} \bar{a_3} \bar{b_1} {b_2} \bar{b_3} \bar{b_4} \bar{m_1} {m_2}$ } \\ \hline

\end{tabular}
\egroup}
\end{table*}


\section{Acknowledgments}
The authors gratefully acknowledge the financial supports from the National Natural Science Foundation of China (No. 61973161, 61991404), Jiangsu Science and technology planning project (No. be2021610).

\bibliographystyle{elsarticle-num} 
\bibliography{draft}

\end{document}